%% file: main.tex
\pdfoutput=1
\PassOptionsToPackage{dvipsnames,table}{xcolor}
\documentclass[]{antgroup}
\PassOptionsToPackage{numbers,sort&compress}{natbib}
\usepackage{antgroup}

\usepackage{graphicx}
\usepackage{multirow}
\usepackage{amsmath}
\usepackage{subcaption}
\usepackage{amssymb}
\usepackage{array}
\usepackage{bm}
\usepackage{enumitem}
\usepackage{tabularx}
\usepackage{booktabs}
\usepackage{threeparttable}
\usepackage{makecell}
\usepackage{colortbl}
\usepackage{xspace}
\usepackage[normalem]{ulem}
\usepackage{caption}
\input{setting-ai}
\usepackage{xcolor}

\setcitestyle{numbers,square,citesep={,},aysep={,},yysep={,}}
\makeatletter
\def\section{\@startsection{section}{1}{\z@}{-2.4ex plus -0.5ex minus -.2ex}{1.2ex plus .2ex}{\sffamily\large\bfseries\color{seedblue}\raggedright}}
\def\subsection{\@startsection{subsection}{2}{\z@}{-1.9ex plus -0.5ex minus -.2ex}{0.8ex plus .2ex}{\sffamily\normalsize\bfseries\color{seedblue}\raggedright}}
\def\@seccntformat#1{\csname the#1\endcsname\hspace{0.7em}}
\makeatother
\renewenvironment{abstract}{%
  \begin{tcolorbox}[enhanced,colback=white,colframe=seedblue!55,boxrule=0.6pt,arc=2.5mm,
    left=7mm,right=7mm,top=3mm,bottom=4mm,before skip=1mm,after skip=3mm]
  {\centering\sffamily\bfseries\large\color{seedblue}Abstract\par}\vspace{2mm}}%
  {\par\vspace{2mm}{\setlength{\parskip}{0pt}\checkdatalist\par}\end{tcolorbox}}
\DeclareCaptionFont{sftitle}{\renewcommand\textbf[1]{{\sffamily\bfseries ##1}}}
\renewcommand{\beginappendix}{\appendix{\sffamily\bfseries\Large\color{seedblue}Appendix\par}\vspace{1ex}}

\makeatletter
\DeclareRobustCommand\onedot{\futurelet\@let@token\@onedot}
\def\@onedot{\ifx\@let@token.\else.\null\fi\xspace}
\def\eg{\emph{e.g}\onedot}

\makeatother
\newcolumntype{C}[1]{>{\centering\arraybackslash}p{#1}}

\newcommand{\method}{\texorpdfstring{\gradtext{Bowerbird}}{Bowerbird}\xspace}
\definecolor{gradA}{HTML}{5A5FD8}\definecolor{gradB}{HTML}{3976F8}\definecolor{gradC}{HTML}{19A7D6}
\makeatletter
\newcount\grad@n \newcount\grad@i \newcount\grad@t
\def\grad@count#1{\ifx#1\relax\else\advance\grad@n\@ne\expandafter\grad@count\fi}
\def\grad@do#1#2{%
  \grad@t=\numexpr\grad@i*100/(\grad@n-1)\relax
  \ifnum\grad@t>45 \colorlet{grad@c}{gradC!\the\numexpr(\grad@t-45)*100/55\relax!gradB}%
  \else\colorlet{grad@c}{gradB!\the\numexpr\grad@t*100/45\relax!gradA}\fi
  {\color{grad@c}#1}%
  \ifx#2\relax\else
    \setbox\z@\hbox{#1#2}\setbox\tw@\hbox{#1}\setbox4\hbox{#2}\kern\dimexpr\wd\z@-\wd\tw@-\wd4\relax
  \fi
  \advance\grad@i\@ne
  \ifx#2\relax\expandafter\@gobble\else\expandafter\grad@do\fi#2}
\DeclareRobustCommand\gradtext[1]{\begingroup\grad@n\z@\grad@i\z@
  \grad@count#1\relax\grad@do#1\relax\relax\endgroup}
\makeatother

\newcommand{\ind}{\mathbb{1}}

\AtBeginDocument{\expandafter\def\expandafter\UrlBreaks\expandafter{\UrlBreaks\do0\do1\do2\do3\do4\do5\do6\do7\do8\do9\do a\do b\do c\do d\do e\do f}}
\definecolor{ourrow}{RGB}{211,243,251}

\definecolor{deltaup}{RGB}{31,122,61}
\definecolor{deltadown}{RGB}{184,67,47}

\definecolor{tablegroup}{RGB}{242,242,242}
\definecolor{tablebest}{RGB}{215,236,255}

\newcommand{\bestscore}[1]{{\bfseries\boldmath #1}}
\newcommand{\secondscore}[1]{\underline{#1}}
\newtheoremstyle{finding}{4pt}{4pt}{\itshape}{}{\bfseries}{.}{.5em}{}
\theoremstyle{finding}
\newtheorem{finding}{Finding}
\crefname{finding}{Finding}{Findings}\Crefname{finding}{Finding}{Findings}
\crefname{figure}{Fig.}{Figs.}\Crefname{figure}{Figure}{Figures}
\crefname{table}{Tab.}{Tabs.}\Crefname{table}{Table}{Tables}
\crefname{equation}{Eq.}{Eqs.}\Crefname{equation}{Equation}{Equations}
\crefname{section}{Sec.}{Secs.}\Crefname{section}{Section}{Sections}
\crefname{appendix}{Sec.}{Secs.}\Crefname{appendix}{Section}{Sections}

\renewcommand\thefootnote{}

\title{\raisebox{-0.15em}{\includegraphics[height=1.15em]{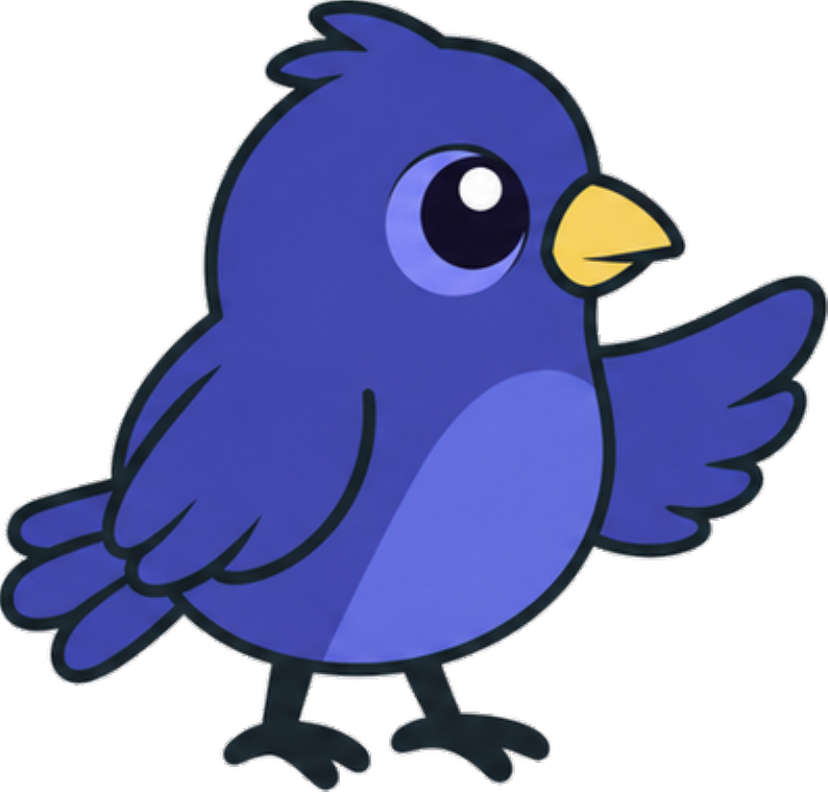}}~\gradtext{Bowerbird}: Evolving and Internalizing Behavioral Strategies for Long-Horizon Visual Coding}

\author{\centering
Hai-Long Sun$^{1,*}$,
Chenglei Wang$^{1,*}$,
Jingdong Chen$^{1,\dag}$,
Han Peng$^{1,\dag}$
}

\footnotetext{$^*$ Equal contribution. $^\dag$ Corresponding authors.}

\affiliation{$^1$Ant Group\\[3pt]{\ttfamily\{yimo.shl, chenglei.wcl, jingdongchen.cjd, penghan.peng\}@antgroup.com}}
\checkdata[Project Page]{\url{https://bowerbird-2026.github.io/}}

\begin{document}
\maketitle

\renewcommand\thefootnote{\arabic{footnote}}
\setcounter{footnote}{0}

\input{sec/0_abstract}
\input{sec/teaser_fig}
\input{sec/1_intro}
\input{sec/2_related}
\input{sec/3_setup}
\input{sec/4_method}
\input{sec/5_experiments}
\input{sec/6_conclusion}

\bibliographystyle{antgroup}
\bibliography{main}

\input{sec/X_suppl}

\end{document}

%% file: setting-ai.tex
\newcommand*\justify{%
  \fontdimen2\font=0.4em%
  \fontdimen3\font=0.2em%
  \fontdimen4\font=0.1em%
  \fontdimen7\font=0.1em%
  \hyphenchar\font=`\-%
}

\renewcommand{\texttt}[1]{%
  \begingroup
  \ttfamily
  \begingroup\lccode`~=`/\lowercase{\endgroup\def~}{/\discretionary{}{}{}}%
  \begingroup\lccode`~=`[\lowercase{\endgroup\def~}{[\discretionary{}{}{}}%
  \begingroup\lccode`~=`.\lowercase{\endgroup\def~}{.\discretionary{}{}{}}%
  \catcode`/=\active\catcode`[=\active\catcode`.=\active
  \justify\scantokens{#1\noexpand}%
  \endgroup
}

\usepackage{amsmath}
\usepackage{amssymb}
\usepackage{amsfonts}
\usepackage{amsthm}
\usepackage[mathcal]{eucal}
\usepackage{mathrsfs}
\usepackage{bm}
\usepackage{blkarray}
\usepackage{nicefrac}

\usepackage{wrapfig}
\usepackage{graphicx}

\usepackage{caption}
\usepackage{cleveref}

\usepackage{tikz}
\usepackage{circuitikz}
\usetikzlibrary{patterns,snakes}
\usetikzlibrary{positioning,calc,fit,decorations.pathmorphing,shapes.geometric, shapes.gates.logic.US, calc}
\usetikzlibrary{arrows,arrows.meta,decorations.markings,shapes,shapes.arrows}
\usetikzlibrary{decorations,decorations.pathreplacing}
\usetikzlibrary{backgrounds}
\usepackage{filecontents}
\usepackage{pgfplots}
\usepackage{pgfplotstable}
\usepgfplotslibrary{groupplots}
\usepackage{scalefnt}
\pgfplotsset{compat=newest}

%% file: sec/0_abstract.tex
\begin{abstract}
Multimodal agents can now develop websites from reference screenshots through long-horizon workflows of perception, coding, and verification.
Existing methods typically train agents through supervised fine-tuning (SFT) on trajectories selected by final task outcomes, or reinforcement learning (RL) with trajectory-level rewards.
However, even successful trajectories can contain redundant actions, regressions, or inadequate verification. Outcome-level supervision provides limited feedback on which intermediate steps to reinforce and which to correct.
In this paper, we propose \method, a framework that combines step-level evaluation, feedback-guided strategy evolution, and strategy internalization.
Specifically, step-level evaluation provides dense progress feedback by rewarding newly passed functional and visual checks, penalizing regressions, and accounting for turn and token costs.
Using this feedback, strategy evolution develops a self-evolving behavioral playbook without updating model parameters, retaining only revisions that improve validation performance.
Strategy internalization supplies the evolved playbook as privileged input during SFT, followed by RL with dense step rewards and no playbook. At inference, the agent requires no additional strategy guidance.
Across challenging visual coding tasks, \method improves the overall score over outcome-based training by 18.8\%, while reducing token usage and interaction turns by 31.8\% and 32.2\%, respectively.
It also achieves state-of-the-art results among models below 500B parameters and even matches closed-source models.
\end{abstract}

%% file: sec/teaser_fig.tex
\begin{figure}[tp]
  \centering
  \includegraphics[width=\linewidth]{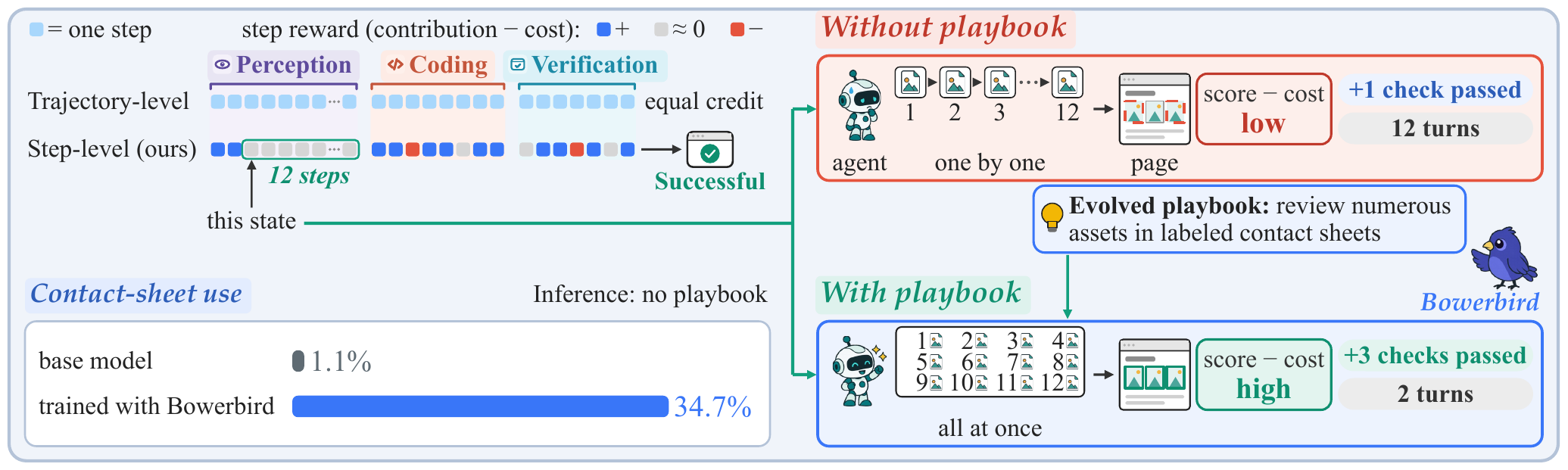}
  \caption{\textbf{From step-level feedback to autonomous strategy use.}
Step-level rewards distinguish useful and wasteful actions within successful trajectories (top left).
The evolved playbook guides more effective execution from the same state (right).
Training boosts unguided strategy use at inference (bottom left).}
  \label{fig:teaser}
\end{figure}

%% file: sec/1_intro.tex
\section{Introduction}
\label{sec:intro}

Agents built on multimodal large language models (MLLMs) can help users develop web applications using code editors, browsers, and testing tools~\cite{yang2024sweagent,wang2024codeact,bai2025qwen3vl}.
Given reference screenshots and functional requirements, they interpret designs, write code, and iteratively refine applications by running websites, inspecting results, and making successive edits.
As tasks expand from static webpages to interactive and full-stack websites~\cite{beltramelli2018pix2code,laurencon2024websight,si2025design2code,xiao2025interaction2code,he2026vision2web}, it becomes necessary to meet both visual and functional requirements.
Despite recent advances in multimodal coding agents, reliably and efficiently developing complex websites remains challenging.
Such tasks can span tens to hundreds of interaction turns, with website quality and development cost depending on which assets agents inspect, which problems they prioritize, when they verify their work, and when they stop.

Existing methods typically train agents through supervised fine-tuning (SFT) on trajectories selected by final outcomes~\cite{chen2023fireact,zeng2024agenttuning,pan2025swegym,yang2025swesmith} or reinforcement learning (RL) with trajectory-level rewards~\cite{wei2025swerl,golubev2025multiturnswe,luo2025deepswe,kimiteam2025k2}.
However, final outcomes provide limited feedback on intermediate actions: even successful trajectories may repeatedly rewrite stylesheets, break working pages, or omit verification after the last edit (\cref{fig:teaser}).
These observations motivate us to look beyond final outcomes and examine how agents select and apply strategies.
Therefore, this raises a natural question: \noindent\emph{Do agents lack the ability to execute effective strategies, or do they fail to use them when needed?}

We find that the base model can exhibit useful development behaviors but rarely adopts them spontaneously.
On the evaluation tasks, for example, it organizes image assets into a labeled contact sheet before coding in only 1.1\% of its trajectories and inspects the rendered page after its last edit in only 10.6\%.
These empirical observations suggest a gap between demonstrated capabilities and spontaneous strategy adoption.

To bridge this gap, we propose \method, a training framework combining stage-aware step-level evaluation, feedback-guided strategy evolution, and strategy internalization.
First, \emph{step-level evaluation} assigns action-level credit based on information value during perception, code progress during coding, and changes in passed checks during verification, while accounting for turn and token costs.
This feedback distinguishes productive actions from redundant or harmful ones across development stages.
Building on these signals, \emph{strategy evolution} develops a self-evolving behavioral playbook without updating model parameters.
Starting from an empty playbook, it iteratively refines strategies and retains only revisions that improve validation performance.
Finally, \emph{strategy internalization} incorporates the evolved strategies into the agent through SFT followed by RL.
The playbook guides trajectory regeneration and serves as privileged input during SFT, providing demonstrations of strategy execution.
We then remove the playbook during RL and use dense step rewards to refine when to apply strategies and when to stop.
Together, these stages enable the agent to apply the learned strategies without additional playbook guidance at inference.

As shown in \cref{tab:main}, \method achieves an overall score of 78.2 on the visual coding benchmark, outperforming outcome-based training by 12.4 points (18.8\%).
Compared with the same baseline, \method reduces token usage and interaction turns by 31.8\% and 32.2\%, respectively.
Notably, it also outperforms all evaluated open-source models below 500B total parameters and selected closed-source models, while remaining competitive on general multimodal benchmarks.

In summary, \method offers
\textbf{1)} empirical evidence of limited spontaneous strategy adoption, together with stage-aware step-level evaluation that accounts for progress and execution costs;
\textbf{2)} feedback-guided strategy evolution and internalization that enable agents to learn and apply reusable strategies without inference-time playbook guidance; and
\textbf{3)} improved visual coding performance and execution efficiency, alongside competitive performance on general multimodal benchmarks.

%% file: sec/2_related.tex
\section{Related Work}
\label{sec:related}

\noindent\textbf{Visual coding.}
Research on website generation spans screenshot-to-code translation~\cite{beltramelli2018pix2code}, webpage generation~\cite{laurencon2024websight,si2025design2code}, long and interactive pages~\cite{longwebbench2026,xiao2025interaction2code}, and text-driven or full-stack development~\cite{lu2025webgenbench,he2026vision2web}.
Prior work evaluates outputs through visual similarity~\cite{jiang2025screencoder,chen2025rrvf}, page quality scores~\cite{xiao2025aescoder,gao2026aescode,webgenr12026}, or stepwise review and interaction tests~\cite{lu2026webgenagent,relook2025,rila2026}.
However, feedback from rendered pages provides little direct supervision for perception and coding steps taken before the website can run.
In this work, we assign credit to these steps by estimating the value of acquired information and measuring implementation progress.

\noindent\textbf{From outcome-level to step-level supervision.}
Agents are commonly trained by imitating successful trajectories~\cite{zelikman2022star,yuan2023rft,chen2023fireact,zeng2024agenttuning,pan2025swegym,yang2025swesmith} or using final scores as RL rewards~\cite{wei2025swerl,golubev2025multiturnswe,luo2025deepswe,kimiteam2025k2,deepseekai2025v32}.
With outcome-based group-relative rewards, actions within a trajectory share the same advantage~\cite{shao2024deepseekmath}.
Other approaches learn from failures and recoveries~\cite{song2024eto,yuan2025agentr}, select efficient trajectories~\cite{han2026swetrace,li2026tqs}, or mask steps judged unhelpful by an LLM~\cite{sweprime2026,slinko2026srft}.
Process reward models~\cite{lightman2024verify,wang2024mathshepherd,choudhury2025agentprm,xi2025agentprm,he2026procua} and step-level agentic RL~\cite{wei2025turnlevel,feng2025gigpo,wang2025sparl,ding2026proas,zhang2026trca,tan2026hcapo,zhou2025sweetrl,he2026workrover} provide finer-grained feedback on individual steps.
Step-level scores evaluate actions within a trajectory, but do not by themselves identify strategies worth reusing across tasks.
To address this, \method compares high- and low-scoring trajectory segments from similar starting states to derive reusable rules, then uses these rules to generate demonstrations of effective behaviors that the agent rarely adopts.

\noindent\textbf{Strategy learning and internalization.}
Prompt optimization and experiential learning derive natural-language guidance through score-driven search~\cite{yang2024opro}, trajectory reflection~\cite{agrawal2026gepa}, or knowledge accumulation~\cite{zhao2024expel,zhang2025ace,cai2025tfgrpo}.
These approaches typically keep the guidance in context at inference, whereas context distillation~\cite{snell2022context,didolkar2025metacognitive} and learning using privileged information~\cite{vapnik2009lupi,privproc2026} incorporate it into training.
However, imitating guided behavior remains insufficient for determining when to apply a strategy.
In this paper, we select playbook revisions based on whole-task performance and execution cost, then use the validated playbook as privileged input during SFT.
During RL, we withdraw the playbook and use step-level rewards to train the agent to decide when to apply a strategy and when to stop.

%% file: sec/3_setup.tex
\section{A Closer Look at Outcome Supervision}
\label{sec:setup}

Multimodal agents combine visual understanding with tool use to complete complex tasks, but final outcomes reveal little about the effectiveness of their intermediate actions.
This issue is particularly pronounced in visual coding, where agents repeatedly write, run, and revise websites from design screenshots, assets, and functional requirements.
These actions may advance the task, repeat earlier work, or undo previous progress.
We therefore study visual coding to examine the relationship between final outcomes and execution behavior.

\subsection{Problem Formulation}
\label{sec:formulation}

\begin{figure}[tp]
\centering
\includegraphics[width=0.88\linewidth]{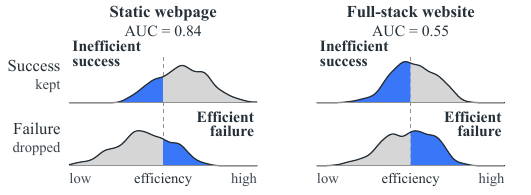}
\caption{\textbf{Final outcomes and execution efficiency can diverge.}
Execution efficiency distinguishes successful from failed trajectories on static webpages (AUC = 0.84), but barely on full-stack websites (AUC = 0.55).
Outcome filtering therefore retains inefficient successes and discards efficient failures.}
\label{fig:decoupling}
\end{figure}

Visual coding covers static webpages, interactive websites, and full-stack websites.
We define each assistant turn as one step: at step $t$, the agent takes action $a_t$ given the history $h_t$ (the task and all preceding actions and observations) and leaves workspace $s_t$; a trajectory $\tau$ has $T$ steps.
The first application-code edit and the first subsequent website rendering or test divide the trajectory into perception, coding, and verification.
Evaluation uses visual and functional checks.
Visual checks compare corresponding regions of the design and rendered page and apply to all tasks.
Functional checks apply only to interactive and full-stack websites: they replay the task-specified workflows in a browser and verify whether each expected outcome occurs.
Each check returns $v_k(s)\in\{0,1\}$, and the website score is the fraction of checks passed, $\Phi(s)=\frac{1}{K}\sum_{k=1}^{K}v_k(s)$, where $K$ is the number of checks.
The pass rates within each check type measure visual and functional performance.
Websites that cannot be built or started fail all checks.

\subsection{Pilot Study}
\label{sec:pilot}\label{sec:pilot_setup}\label{sec:pilot_measure}
To analyze how multimodal models behave during visual coding, we study Qwen3.8-Max~\cite{qwenteam2026qwen38}, a frontier model, on static, interactive, and full-stack tasks using the Claude Code harness.
Given the complexity of these tasks, the resulting trajectories often span more than a hundred interaction turns and can exhibit different patterns of execution.
To further analyze behavioral differences between successful and failed attempts, we classify trajectories into two groups according to whether the delivered websites meet the tasks' visual and functional requirements.
Then, we replay the edit logs to evaluate website states at the end of coding and after each subsequent edit.
By comparing execution efficiency across the two groups, the effects of the same operation across stages, and the base model's behavior with and without strategy guidance, we obtain the following findings, with the data and measurement detailed in Supp.~\cref{sec:supp_pilot}.

\phantomsection\label{sec:pilot_findings}
\begin{finding}
\label{find:decoupling}
Successful delivery does not imply efficient execution, especially on long-horizon tasks.
\end{finding}
An edit is considered productive if it passes at least one new check without breaking previously passed checks or increasing runtime errors.
We define execution efficiency as the fraction of verification edits that are productive and use AUC to measure how well it distinguishes successful from failed trajectories. Values closer to 0.5 indicate weaker discrimination.
As shown in \cref{fig:decoupling}, AUC decreases from 0.84 on static tasks to 0.55 on full-stack tasks, approaching the chance level of 0.5.
Moreover, 62.8\% of successful full-stack trajectories contain edits that break previously passed checks.
Outcome-based trajectory selection therefore retains successful trajectories despite inefficient edits and regressions.
This motivates us to evaluate the contributions of individual actions beyond final task outcomes.

\begin{finding}
\label{find:frequency}
The right action at the wrong time is still wrong.
\end{finding}
We compare the mean step rewards of the same operation types during coding and verification.
Each step reward measures an action's estimated contribution minus its execution costs (\cref{eq:utility}).
For some common operations, the mean rewards have opposite signs across the two stages.
For example, whole-file rewrites have a positive mean reward during coding but a negative one during verification.
In the latter stage, 31.6\% of rewrites break previously passed checks.
This motivates us to learn strategies that specify both which operations to use and when to use them.

\begin{finding}
\label{find:adoption}
Models can execute effective strategies but seldom use them autonomously.
\end{finding}
Reviewing assets in contact sheets and cropping missing assets receive high mean step rewards when both action contributions and execution costs are considered.
However, these strategies occur in only 3.8\% and 1.6\% of successful trajectories generated by Qwen3.8-Max, respectively.
To further examine our own base model's ability to execute these strategies, we provide explicit instructions on when to use each.
Among tasks where both strategies apply, the proportion of trajectories using both strategies rises from 4.8\% without guidance to 63.9\% with guidance.
This demonstrates that the model can execute both strategies when prompted, but rarely uses them on its own.

%% file: sec/4_method.tex
\section{The \method Framework}
\label{sec:method}

\begin{figure*}[tp]
  \centering
  \includegraphics[width=1.0\linewidth]{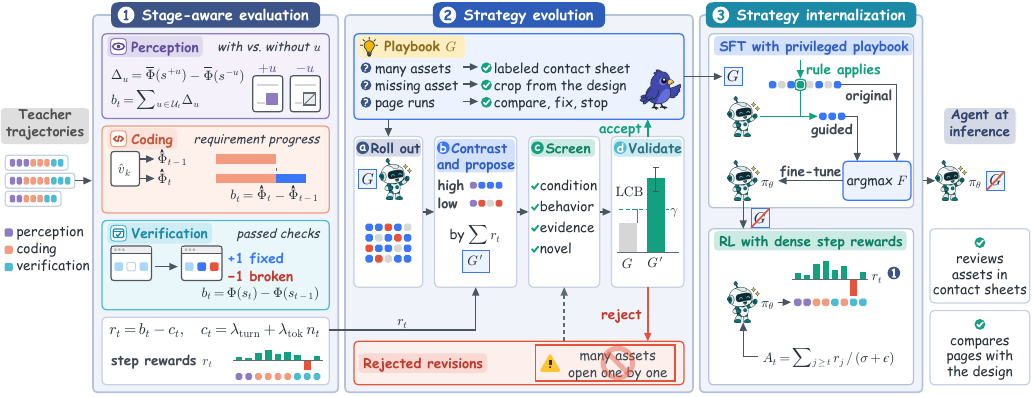}
  \caption{\textbf{Overview of \method.}
  Stage-aware step-level evaluation estimates action contributions and execution costs.
  Rule revisions from contrasting trajectory segments are validated on complete tasks to build a behavioral playbook.
  The playbook guides trajectory generation and SFT and is removed during RL, where step rewards refine strategy use.}
  \label{fig:framework}
\end{figure*}

We observe that successful outcomes can conceal inefficient execution, action effectiveness varies across stages, and models can execute effective strategies that they rarely adopt on their own.
Building on these findings, \method evaluates action contributions within each stage, extracts and validates rules from contrasting trajectories, and uses SFT followed by RL to teach strategy execution and refine when strategies are applied (\cref{fig:framework}).

\subsection{Stage-Aware Step-Level Evaluation}
\label{sec:attr}

Runtime checks measure progress only once the website is built and running, making it difficult to assign credit to earlier perception and coding actions.
To evaluate actions throughout the trajectory, we estimate their contributions from stage-specific evidence and subtract turn and token costs to obtain the step reward:
\begin{equation}
  r_t=b_t-c_t,\qquad
  c_t=\lambda_{\mathrm{turn}}+\lambda_{\mathrm{tok}}n_t.
  \label{eq:utility}
\end{equation}
The contribution $b_t$ measures the action's benefit, $c_t$ is its execution cost, $n_t$ counts generated tokens, and $\lambda_{\mathrm{turn}}$ and $\lambda_{\mathrm{tok}}$ control turn and token costs.
Stages and the website score $\Phi(s)$ follow \cref{sec:formulation}.

To compare action contributions across stages, we use the normalized website score as a common reference.
We divide the development process into three broad stages: perception, coding, and verification.
For these stages, we measure the effect of acquired information on the final score, progress in requirement implementation, and changes in passed checks, respectively.

\noindent\textbf{Perception.}
Design inspection and asset preparation affect website quality through subsequent development.
We measure their contribution by comparing final scores with and without an information unit $u$.
A unit is an observation of an asset or design region, or a prepared asset file.
Starting from the history $h_{\mathrm P}$ and workspace $s_{\mathrm P}$ at the end of perception, we use a fixed reference policy $\pi_{\mathrm{ref}}$ to sample $Q$ continuations under each condition, indexed by $q$.
Masking, denoted $\setminus u$, removes the corresponding observations and files and blocks subsequent access to $u$:
\begin{equation}
  \begin{aligned}
    s_q^{+u}
    &\sim\pi_{\mathrm{ref}}(\cdot\mid h_{\mathrm P},s_{\mathrm P}),\\
    s_q^{-u}
    &\sim\pi_{\mathrm{ref}}
    (\cdot\mid h_{\mathrm P}\setminus u,\,s_{\mathrm P}\setminus u),\\
    \Delta_u
    &=\frac{1}{Q}\sum_{q=1}^{Q}
    \bigl[\Phi(s_q^{+u})-\Phi(s_q^{-u})\bigr].
  \end{aligned}
  \label{eq:information}
\end{equation}
The contribution at step $t$ is
$b_t=\sum_{u\in\mathcal U_t}\Delta_u$,
where $\mathcal U_t$ contains only units first observed or prepared at that step, avoiding repeated credit.
For efficiency, we estimate information values on one reference trajectory per task and reuse them for other trajectories, assigning zero to units not covered by the reference.

\noindent\textbf{Coding.}
During coding, we assess progress by how completely the current implementation satisfies the task requirements.
A fixed MLLM evaluator, described in Supp.~\cref{sec:supp_evaluator}, estimates the implementation progress $\hat v_k(h_t,s_t)\in[0,1]$ of each of the $K$ requirements before and after each edit:
\begin{equation}
  \begin{aligned}
    \hat\Phi_t
    &=\frac{1}{K}\sum_{k=1}^{K}
    \hat v_k(h_t,s_t),\\
    b_t
    &=\hat\Phi_t-\hat\Phi_{t-1}.
  \end{aligned}
  \label{eq:coding}
\end{equation}
The evaluator uses the task's check requirements, current workspace code, and textual and visual observations in the execution history.
Each $\hat v_k$ is a soft counterpart of the check result $v_k$.
Their mean, $\hat\Phi_t$, summarizes overall implementation progress, and its change across an edit gives the coding contribution.
For example, steps that only read source files or search the codebase leave the implementation unchanged and therefore receive zero direct contribution ($b_t=0$), while still incurring execution costs.
The last coding edit yields the first rendered workspace $s_{\mathrm C}$, so its estimate is replaced by the measured score, $b_t=\Phi(s_{\mathrm C})-\hat\Phi_{t-1}$, preventing reward hacking via code that the evaluator overrates.

\noindent\textbf{Verification.}
Once the website is runnable, we evaluate each edit through changes in checks passed by the checkers of Supp.~\cref{sec:supp_checkers}:
\begin{equation}
  \begin{aligned}
    b_t
    &=\Phi(s_t)-\Phi(s_{t-1})\\
    &=\frac{1}{K}\sum_{k=1}^{K}
    \bigl[v_k(s_t)-v_k(s_{t-1})\bigr].
  \end{aligned}
  \label{eq:verify}
\end{equation}
Newly passed checks increase $b_t$, while failures of previously passed checks reduce it, including regressions that produce no runtime errors.
Steps that only inspect the page receive zero direct contribution, $b_t=0$, while their downstream benefits are captured by cumulative RL advantages.

Together, these stage-specific contribution estimates and execution costs help distinguish productive actions from unproductive repetition and regressions, providing step-level feedback for subsequent strategy learning.

\subsection{Feedback-Guided Strategy Evolution}
\label{sec:evolve}
Locally effective behaviors require validation on complete tasks before being retained as reusable rules.
Starting from an empty playbook $G$, we evolve rules of the form
``if \emph{condition}, then \emph{behavior}''
without updating model parameters.
At each iteration, the teacher solves tasks with the current playbook and compares trajectory segments with similar starting situations but different cumulative step rewards.
Each segment extends to the end of its stage to capture downstream effects within that stage.
Each comparison-guided revision adjusts at most $m$ rules and yields a candidate playbook $G'$.

\noindent\textbf{Proposal screening.}
Each proposal must include an applicability condition, an executable behavior, and supporting trajectory evidence.
Proposals that omit conditions, restate task-specific answers, or duplicate existing rules are rejected.
Conflicts are resolved by refining conditions or revising existing rules.

\noindent\textbf{Task validation.}
To measure the effect of a revision on complete tasks, we define trajectory utility as the final website score minus total execution cost,
$F(\tau)=\Phi(s_T)-\sum_{t=1}^{T}c_t$.
Candidates first undergo small-batch task testing, and those retained are evaluated on a held-out validation set $\mathcal D_{\mathrm{val}}$ using the mean utility:
\begin{equation}
  J(G)=\frac{1}{|\mathcal D_{\mathrm{val}}|}
  \sum_{x\in\mathcal D_{\mathrm{val}}}F(\tau_x^G),
  \label{eq:validation}
\end{equation}
where $\tau_x^G$ is the trajectory generated for task $x$ with playbook $G$.
We compare the current playbook $G$ and the candidate $G'$ on the same validation tasks, accepting the revision only when:
\begin{equation}
  \operatorname{LCB}_{0.95}\bigl(
    J(G')-J(G)
  \bigr)>\gamma.
  \label{eq:deficit}
\end{equation}
Here, $\operatorname{LCB}_{0.95}$ denotes the 95\% lower confidence bound estimated by paired bootstrap over task-wise utility differences, and $\gamma$ specifies the minimum required improvement.
An accepted revision replaces $G$ with $G'$; otherwise, the current playbook is retained.
Rejected revisions are recorded and are not retested without substantive changes.
Evolution stops after several consecutive iterations without acceptance; Supp.~\cref{sec:supp_guidance} lists all revisions and the final playbook.
The tasks used for evolution, validation, and final evaluation are mutually disjoint, and their construction is described in Supp.~\cref{sec:supp_pipeline}.

Feedback-guided strategy evolution extracts rules from contrasting trajectories and retains revisions that improve whole-task utility, building a playbook for trajectory generation and strategy internalization.

\subsection{Strategy Internalization from SFT to RL}
\label{sec:inject}
Effective strategies are infrequent in the original trajectories, so filtering alone yields few demonstrations of how to execute them.
Strategy internalization combines SFT on rule-guided demonstrations with RL on the model's own rollouts, enabling it to apply learned strategies without playbook guidance at inference.

\noindent\textbf{SFT with a privileged playbook.}
\phantomsection\label{sec:sft}
The teacher uses $G$ to regenerate continuations from training states where a rule applies.
For each task $x$, we retain the trajectory with the highest cumulative step reward $\sum_t r_t$ among the original and guided regenerations, selecting demonstrations by final website quality, the information value of perception, and execution cost.
The selected trajectories form $\mathcal D_{\mathrm{SFT}}$, and the student policy is trained with:\looseness=-1
\begin{equation}
  \mathcal L_{\mathrm{SFT}}(\theta)
  =-\mathbb E_{\tau\sim\mathcal D_{\mathrm{SFT}}}
  \left[
    \sum_{t=1}^{T}
    \log\pi_\theta(a_t\mid h_t,G)
  \right].
  \label{eq:sft}
\end{equation}
Here, $\pi_\theta$ is the trainable policy, which predicts action $a_t$ from $h_t$ and $G$, with the playbook serving as privileged input during SFT.

\noindent\textbf{RL with dense step rewards.}
Imitation does not directly determine when to use strategies and may retain redundant actions.
After SFT, we remove $G$ and collect rollouts from the model itself.
Dense step rewards refine strategy use and stopping through cumulative step advantages:
\begin{equation}
  A_t^{(i)}
  =\sum_{j=t}^{T_i}\frac{r_j^{(i)}}{\sigma+\epsilon}.
  \label{eq:rl_reward}
\end{equation}
For rollout $i$ of length $T_i$, $\sigma$ is the standard deviation of step rewards across the $M$ rollouts for the same task, and $\epsilon$ ensures numerical stability.
All tokens in step $t$ share $A_t^{(i)}$.
Cumulative rewards credit earlier preparation and inspection steps for later progress.
Rewards are scaled without mean subtraction, so the cumulative return retains the constant turn penalty.

%% file: sec/5_experiments.tex
\section{Experiments}
\label{sec:exp}

\begin{table*}[tp]
  \centering
  \caption{\textbf{Main results on visual coding tasks.}
\method outperforms outcome-based training across static, interactive, and full-stack tasks, raising the overall score by 12.4 points.
It also achieves the highest interactive score among the evaluated open-source models.
Params: total parameters (--: undisclosed). VS and FS: visual and functional scores.}
  \label{tab:main}
  \setlength{\tabcolsep}{4pt}
  \renewcommand{\arraystretch}{1.0}
  \resizebox{\textwidth}{!}{%
    \begin{tabular}{lcc*{11}{C{1.25cm}}}
      \toprule
      \multirow{2}{*}{\textbf{Model}}
        & \multirow{2}{*}{\textbf{Params}}
        & \multirow{2}{*}{\textbf{Harness}}
        & \multirow{2}{*}{\textbf{Overall}}
        & \multicolumn{4}{c}{\textbf{STATIC WEBPAGE}}
        & \multicolumn{3}{c}{\textbf{INTERACTIVE}}
        & \multicolumn{3}{c}{\textbf{FULL-STACK}} \\
      \cmidrule(lr){5-8}
      \cmidrule(lr){9-11}
      \cmidrule(lr){12-14}
       & & & & Desktop & Tablet & Mobile & Avg
           & VS & FS & Avg
           & VS & FS & Avg \\
      \midrule
      \rowcolor{tablegroup}
      \multicolumn{14}{c}{\itshape Open-Source Models} \\
      Kimi-K3~{\scriptsize\cite{kimiteam2026k3}}
        & 2.8T & Claude Code & 79.0
        & 85.1 & 80.4 & 78.8 & 81.4
        & 71.1 & 92.1 & 81.6
        & 65.2 & 82.8 & 74.0 \\
      GLM-5.3-Flash~{\scriptsize\cite{zai2026glm53flash}}
        & 320B & Claude Code & 77.8
        & 89.6 & 85.1 & 85.4 & 86.7
        & 68.5 & 89.1 & 78.8
        & 58.7 & 77.4 & 68.0 \\
      DeepSeek-V4.1-Flash~{\scriptsize\cite{deepseekai2026v41flash}}
        & 552B & Claude Code & 72.6
        & 81.2 & 78.6 & 77.9 & 79.2
        & 59.3 & 90.4 & 74.9
        & 49.5 & 78.0 & 63.8 \\
      DeepSeek-V4-Flash-Vision~{\scriptsize\cite{deepseekai2026v4flashvision}}
        & 305B & Claude Code & 67.6
        & 74.8 & 69.5 & 67.3 & 70.5
        & 53.2 & 86.9 & 70.0
        & 45.8 & 78.4 & 62.1 \\
      MiniMax-M3~{\scriptsize\cite{minimax2026m3}}
        & 428B & Claude Code & 59.0
        & 58.8 & 51.6 & 51.8 & 54.1
        & 47.9 & 83.4 & 65.7
        & 42.4 & 72.4 & 57.4 \\
      Kimi-K2.6~{\scriptsize\cite{moonshot2026kimik26}}
        & 1T & Claude Code & 54.8
        & 57.3 & 52.7 & 50.8 & 53.6
        & 40.1 & 81.2 & 60.6
        & 36.2 & 64.2 & 50.2 \\
      Qwen3.5-122B-A10B~{\scriptsize\cite{qwenteam2026qwen35}}
        & 122B & Claude Code & 44.9
        & 69.7 & 62.7 & 58.0 & 63.5
        & 24.7 & 56.6 & 40.7
        & 19.4 & 41.9 & 30.7 \\
      Qwen3-VL-8B~{\scriptsize\cite{bai2025qwen3vl}}
        & 8B & Claude Code & 4.0
        & 12.4 & 11.9 & 11.3 & 11.9
        & 0.0 & 0.0 & 0.0
        & 0.0 & 0.0 & 0.0 \\
      \midrule
      \rowcolor{tablegroup}
      \multicolumn{14}{c}{\itshape Closed-Source Models} \\
      GPT-6-Astra~{\scriptsize\cite{openai2026gpt6astra}}
        & -- & Codex & 86.5
        & 94.4 & 93.1 & 93.4 & 93.6
        & 82.6 & 92.8 & 87.7
        & 75.1 & 81.0 & 78.0 \\
      Claude-Fable-5~{\scriptsize\cite{anthropic2026fable5}}
        & -- & Claude Code & 82.3
        & 89.6 & 86.7 & 86.1 & 87.4
        & 78.1 & 88.3 & 83.2
        & 69.0 & 83.6 & 76.3 \\
      Claude-Opus-5~{\scriptsize\cite{anthropic2026opus5}}
        & -- & Claude Code & 80.2
        & 91.2 & 89.5 & 89.8 & 90.1
        & 74.4 & 88.7 & 81.5
        & 68.0 & 70.0 & 69.0 \\
      Gemini-3.7-Flash~{\scriptsize\cite{google2026gemini37flash}}
        & -- & Gemini CLI & 76.5
        & 86.4 & 81.6 & 80.3 & 82.8
        & 63.1 & 86.7 & 74.9
        & 57.0 & 86.6 & 71.8 \\
      Claude-Opus-4.8~{\scriptsize\cite{anthropic2026opus48}}
        & -- & Claude Code & 76.1
        & 82.0 & 74.1 & 70.9 & 75.7
        & 67.9 & 91.1 & 79.5
        & 60.6 & 85.7 & 73.1 \\
      GPT-6-Luna~{\scriptsize\cite{openai2026gpt6luna}}
        & -- & Codex & 76.0
        & 89.2 & 85.0 & 87.7 & 87.3
        & 60.2 & 95.9 & 78.0
        & 51.3 & 73.9 & 62.6 \\
      Qwen3.8-Max~{\scriptsize\cite{qwenteam2026qwen38}}
        & 2.4T & Claude Code & 74.7
        & 77.6 & 71.4 & 69.7 & 72.9
        & 62.2 & 91.2 & 76.7
        & 60.3 & 88.4 & 74.3 \\
      GPT-5.6-Sol~{\scriptsize\cite{openai2026gpt56}}
        & -- & Codex & 73.5
        & 87.9 & 84.4 & 85.0 & 85.7
        & 64.8 & 83.2 & 74.0
        & 49.1 & 72.1 & 60.6 \\
      Claude-Opus-4.6~{\scriptsize\cite{anthropic2026opus46}}
        & -- & Claude Code & 57.6
        & 60.4 & 54.7 & 54.6 & 56.6
        & 44.5 & 75.2 & 59.8
        & 41.9 & 71.1 & 56.5 \\
      Gemini-3.1-Pro-Preview~{\scriptsize\cite{google2026gemini31pro}}
        & -- & Claude Code & 28.8
        & 47.7 & 40.5 & 41.1 & 43.1
        & 21.4 & 37.8 & 29.6
        & 10.3 & 17.0 & 13.6 \\
      Doubao-Seed-2.0-Pro~{\scriptsize\cite{bytedanceseed2026seed2}}
        & -- & Claude Code & 21.5
        & 40.7 & 34.7 & 31.2 & 35.5
        & 18.5 & 33.5 & 26.0
        & 3.5 & 2.5 & 3.0 \\
      \midrule
      \rowcolor{tablegroup}
      \multicolumn{14}{c}{\itshape Our Models} \\
      Our Base Model
        & 124B & Claude Code & 57.7
        & 65.3 & 58.1 & 55.6 & 59.7
        & 47.2 & 84.6 & 65.9
        & 32.2 & 62.8 & 47.5 \\
      \quad + Outcome-based training
        & 124B & Claude Code & 65.8
        & 80.5 & 73.2 & 69.8 & 74.5
        & 57.0 & 89.4 & 73.2
        & 36.6 & 63.0 & 49.8 \\
      \rowcolor{tablebest}
      \textbf{\method}
        & 124B & Claude Code & \bestscore{78.2}
        & \bestscore{86.7} & \bestscore{81.7}
        & \bestscore{76.5} & \bestscore{81.6}
        & \bestscore{75.8} & \bestscore{94.1}
        & \bestscore{85.0}
        & \bestscore{58.9} & \bestscore{77.1}
        & \bestscore{68.0} \\
      \bottomrule
    \end{tabular}%
  }
\end{table*}

\subsection{Experimental Setup}
\label{sec:exp_setup}

\noindent\textbf{Dataset.}
We evaluate \method on visual coding and general multimodal tasks.
Visual coding benchmarks include Vision2Web~\cite{he2026vision2web}, which covers static webpages, interactive websites, and full-stack websites, and Design2Code~\cite{si2025design2code}, which requires reproducing a reference screenshot in a single HTML file.
On Vision2Web, our model runs in Claude Code 2.1.221~\cite{anthropic2026claudecode} with a 256K-token context and automatic compaction.
Following the official evaluation setup, we use Gemini-3.5-Flash~\cite{google2026gemini35flash} as the visual judge and GLM-5V-Turbo~\cite{glmvteam2026glm5vturbo} for browser interaction and functional verification.
Each attempt has a 36,000-second budget, with at most two retries after failure.
Design2Code compares screenshots of rendered pages with the reference designs.
General capability is evaluated on mobile GUI control with AndroidWorld~\cite{rawles2025androidworld}, document parsing with OmniDocBench v1.5~\cite{ouyang2025omnidocbench}, chart reasoning with CharXiv~\cite{wang2024charxiv}, and multimodal reasoning with MMMU-Pro~\cite{yue2025mmmupro}, MathVista~\cite{mathvista}, and the multimodal subset of HLE~\cite{phan2025hle}.
We follow each benchmark's evaluation protocol, detailed in Supp.~\cref{sec:supp_eval}, and use the reasoning questions of CharXiv.

\begin{figure}[tp]
\centering
\includegraphics[width=0.9\linewidth]{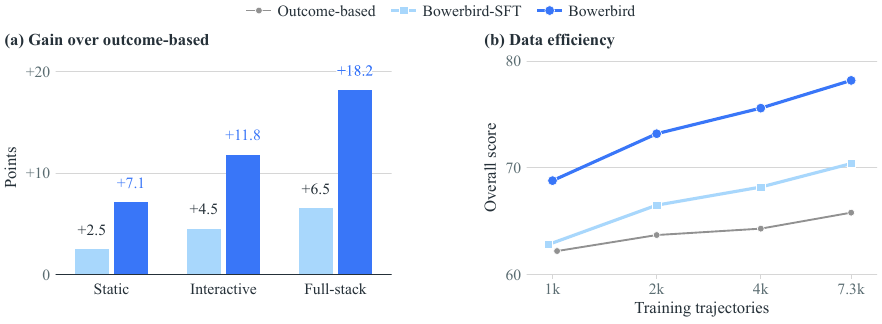}
\caption{\textbf{Data efficiency and gains across task types.}
  (a) Gains of \method and \method-SFT (after SFT) over outcome-based training by task type.
  (b) Overall score versus the number of training trajectories.}
\label{fig:efficiency}
\end{figure}

\noindent\textbf{Implementation Details.}
We use Ling-3.0-flash-VL as the base model.
SFT trains on about 7.3K trajectories (\cref{sec:sft}), using the Muon optimizer with a learning rate of $5\times10^{-5}$, cosine decay, a global batch size of 64, and 400 steps.
The visual encoder and projection layers use 0.1 times the main learning rate.
For RL, we select 200 tasks from the SFT data and sample $M=8$ rollouts per task, with a learning rate of $2\times10^{-5}$, a global batch size of 16, and train 25 steps.
Both SFT and RL are conducted with a context window of 256K tokens.
Training uses BF16 precision on 64 NVIDIA H800 GPUs; other hyperparameters are given in Supp.~\cref{sec:supp_hparams}.

\noindent\textbf{Comparison methods.}
We compare with open-source models from the Kimi, GLM, DeepSeek, MiniMax, and Qwen families, and closed-source models from the GPT, Claude, Gemini, Qwen, and Doubao families.
To examine the effect of training signals, we compare outcome-based training with the step-level training of \method.
The former includes SFT that directly imitates successful trajectories and subsequent RL using the final website score as a trajectory-level reward.
In contrast, \method generates strategy demonstrations under rule guidance and uses step-level rewards to learn when to apply the strategies.

\noindent\textbf{Evaluation metrics.}
Visual coding evaluation reports visual, functional, and overall scores.
Visual scores are obtained by a visual judge comparing rendered screenshots with reference designs, assessing layout, style, and content fidelity.
Functional scores are obtained by a GUI agent executing specified workflows and verifying their expected outcomes.
On Vision2Web, the static score averages desktop, tablet, and mobile scores; interactive and full-stack scores each average visual and functional scores; the overall score averages the three task types.
Design2Code uses its visual reconstruction metric.
For general capability, we report task success rate on AndroidWorld, the overall score combining text, table, and formula parsing on OmniDocBench v1.5, and answer accuracy on CharXiv, MMMU-Pro, MathVista, and the multimodal subset of HLE.\looseness=-1

\subsection{Main Results}
\label{sec:exp_main}
\leavevmode\par\vspace{-0.8\baselineskip}
\begin{wrapfigure}{r}{0.45\linewidth}
\centering
\vspace{-2ex}
\includegraphics[width=\linewidth]{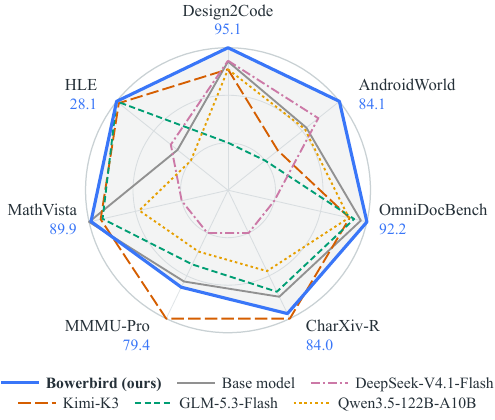}
\caption{\textbf{Transfer and general multimodal performance.}
  Comparison of \method with its base model and open-source models on page reconstruction, mobile GUI control, document parsing, and multimodal reasoning benchmarks.}
\label{fig:general}
\vspace{-2ex}
\end{wrapfigure}
\noindent\textbf{Visual coding performance.}
As shown in \cref{tab:main}, \method raises the overall score of base model from 57.7 to 78.2, achieving 81.6, 85.0, and 68.0 on static, interactive, and full-stack tasks, respectively.
Among the listed open-source models, only Kimi-K3 scores higher overall, while \method achieves the highest interactive score and matches GLM-5.3-Flash on full-stack tasks.
It also outperforms Gemini-3.7-Flash, Claude-Opus-4.8, and GPT-6-Luna overall, and trails only GPT-6-Astra on interactive websites.
Notably, \method exceeds Qwen3.8-Max by 3.5 points with about 5\% of its parameters.
These results show that converting effective strategies into execution demonstrations and refining their use through step-level feedback helps the agent complete long-horizon visual coding tasks.

\begin{table}[tp]
\centering
\caption{\textbf{Ablations of the main components.}
Adding the components cumulatively improves task performance while reducing turns and token usage.}
\label{tab:ablation}
\small
\setlength{\tabcolsep}{8pt}
\begin{tabular}{lrrr}
  \toprule
  Configuration & Overall $\uparrow$ & Turns $\downarrow$ & Tokens $\downarrow$ \\
  \midrule
  Outcome-based training & 65.8 & 107.6 & 119.4K \\
  \hspace*{0.5em}+ Stage-aware trajectory selection & 74.2 & 91.8 & 101.7K \\
  \hspace*{1.0em}+ Evolved-playbook guidance & 77.1 & 80.6 & 90.3K \\
  \rowcolor{tablebest}
  \hspace*{1.5em}+ Step-level RL rewards & \textbf{78.2} & \textbf{72.9} & \textbf{81.4K} \\
  \bottomrule
\end{tabular}
\end{table}

\begin{table}[tp]
\centering
\caption{\textbf{Effect of playbook guidance at inference.}
  We evaluate the same \method model with and without the playbook across all three task types.}
\label{tab:playbook_inference}
\small
\setlength{\tabcolsep}{8pt}
\begin{tabular}{lrrrr}
  \toprule
  Inference & Static & Interactive & Full-stack & Overall \\
  \midrule
  \rowcolor{tablebest}
  Without $G$ & 81.6 & \textbf{85.0} & \textbf{68.0} & \textbf{78.2} \\
  With $G$ & \textbf{83.8} & 81.9 & 66.5 & 77.4 \\
  \bottomrule
\end{tabular}
\end{table}

\noindent\textbf{Data efficiency and scaling.}
To examine how performance scales with training data, we train both methods with different numbers of trajectories.
As shown in \cref{fig:efficiency}b, \method scores 68.8, 73.2, 75.6, and 78.2 with 1K, 2K, 4K, and approximately 7.3K trajectories, exceeding outcome-based training by 6.6, 9.5, 11.3, and 12.4 points, respectively.
With only 1K trajectories, \method already surpasses the baseline trained on all 7.3K trajectories, which scores 65.8.
As the training set grows, both \method's score and its lead over the baseline increase.
These results show that rule-guided trajectory generation and step-level RL improve data efficiency and enable the model to benefit more from additional training data.

\noindent\textbf{Transfer and general capability.}
To assess transfer performance and general multimodal capabilities, we evaluate \method on webpage reconstruction, mobile GUI control, document understanding, and multimodal reasoning benchmarks.
As shown in \cref{fig:general}, \method raises Design2Code and AndroidWorld scores from the base model's 93.5 and 67.2 to 95.1 and 84.1, respectively, surpassing DeepSeek-V4.1-Flash's 93.6 on Design2Code.
It also leads the compared models on OmniDocBench and MathVista, with scores of 92.2 and 89.9, respectively.
On HLE, it matches GLM-5.3-Flash, while ranking second behind Kimi-K3 on both CharXiv reasoning and MMMU-Pro.
These results show that strategy learning for visual coding also leaves the model competitive on document understanding and multimodal reasoning.

\begin{figure*}[tp]
  \centering
  \includegraphics[width=1.0\linewidth]{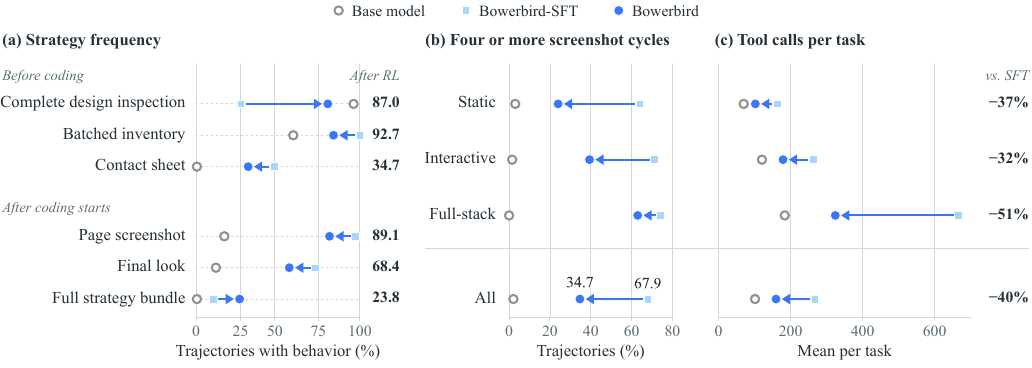}
  \caption{\textbf{Effects of training on strategy use and execution costs.}
  We compare the base model, \method-SFT, and \method without playbook guidance at inference.
  RL applies complete design inspection selectively while reducing repeated verification and tool calls.
  }
  \label{fig:behavior}
\end{figure*}

\subsection{Ablation Study}
\label{sec:exp_ablation}

We conduct cumulative ablations on Vision2Web to examine how trajectory selection, evolved-playbook guidance, and RL rewards affect task quality and execution efficiency.
All configurations use about 7.3K training trajectories, equal SFT and RL steps, and identical evaluation settings.

\noindent\textbf{Comparison with outcome-based training.}
\method raises the overall score from 65.8 to 78.2, while reducing generated tokens from 119.4K to 81.4K (31.8\%) and interaction turns from 107.6 to 72.9 (32.2\%).
As shown in \cref{fig:efficiency}a, gains reach 7.1, 11.8, and 18.2 points on static, interactive, and full-stack tasks, respectively.
On full-stack tasks, the visual score rises from 36.6 to 58.9 and the functional score from 63.0 to 77.1, showing improvements in both visual fidelity and functionality.
The growing advantage on tasks requiring more successive decisions and edits shows the value of rule-guided demonstrations and step-level feedback for long-horizon development.

\noindent\textbf{Component contributions.}
Starting from outcome-based training, we first use stage-aware step rewards to select SFT trajectories while retaining the outcome-based RL reward.
As shown in \cref{tab:ablation}, this raises the overall score from 65.8 to 74.2 and reduces interaction turns and generated tokens to 91.8 and 101.7K, respectively, showing the value of accounting for action contributions and execution costs when selecting training demonstrations.
We then use the evolved playbook $G$ to regenerate demonstrations and include it as input during SFT.
These changes yield a further 2.9-point gain, with turns and tokens decreasing to 80.6 and 90.3K, supporting the use of validated rules to guide both demonstration generation and supervised training.
Finally, keeping the SFT data and setup unchanged, we replace the outcome-based RL reward with step-level rewards.
The score increases to 78.2, while turns and tokens fall to 72.9 and 81.4K.
This supports refining strategy use through contribution and cost feedback after SFT.

\subsection{Further Analysis}
\label{sec:exp_analysis}

\noindent\textbf{Strategy use and execution efficiency.}
To examine how strategy demonstrations and RL change execution behavior, we compare the base model, \method-SFT, and the final \method model, all without playbook guidance.
\Cref{fig:behavior} summarizes their strategy use, repeated verification, and tool calls, with full rates in Supp.~\cref{tab:behavior}.
The base model inspects every complete design image before coding in 97.9\% of trajectories.
After SFT, batched inventory and contact-sheet use rise to 100.0\% and 49.2\%, but the model views designs in slices in 79.8\% of trajectories and inspects every complete design image in only 25.4\%.
After RL, complete design inspection occurs in 87.0\% of trajectories, while slice viewing and contact-sheet use fall to 48.2\% and 34.7\% and batched inventory stays high at 92.7\%.
RL also reduces the proportion of trajectories with four or more screenshot cycles from 67.9\% to 34.7\% and average tool calls from 267.1 to 159.4.
During RL, the overall score rises from 70.4 to 78.2.
These changes show that RL teaches the model when to apply each strategy, performing complete design inspection selectively while reducing repeated verification and tool use.
Cumulative step rewards credit preparation and inspection through downstream progress, while execution costs discourage unproductive repetition.

\noindent\textbf{Effect of playbook guidance at inference.}
The playbook $G$ guides trajectory generation and SFT but is removed during RL and default inference.
To test whether the model still benefits from explicit guidance after strategy internalization, we reintroduce $G$ at evaluation.
As shown in \cref{tab:playbook_inference}, adding $G$ raises the static score by 2.2 points but lowers interactive and full-stack scores by 3.1 and 1.5 points, respectively, reducing the overall score from 78.2 to 77.4.
Explicit rules can interfere with the model's use of learned strategies when they do not fit the current task or execution state.
Without $G$, the model instead chooses when and how to apply these strategies as the task progresses.
These findings validate the effectiveness of strategy internalization.\looseness=-1

%% file: sec/6_conclusion.tex
\section{Conclusion}
\label{sec:conclusion}

In this paper, we present \method, a framework for teaching multimodal agents when and how to use effective strategies in long-horizon visual coding.
It combines stage-aware step-level evaluation, feedback-guided strategy evolution, and strategy internalization to learn from intermediate progress and execution experience.
The evolved playbook guides trajectory generation and SFT, while step-level RL refines strategy use and discourages redundant actions through cost feedback.
The trained model applies these strategies autonomously without playbook guidance at inference.
Experiments show that \method improves visual coding quality and execution efficiency, with competitive results on transfer tasks and general multimodal benchmarks.\looseness=-1

%% file: sec/X_suppl.tex
\clearpage
\beginappendix
\renewcommand{\arraystretch}{1.12}

\section{Teacher Trajectories}
\label{sec:supp_agents}

The trajectory pool is produced by Qwen3.8-Max, the model of the pilot study (\cref{sec:pilot}), a frozen 2.4T-parameter MoE model (95B active) in the Claude Code harness, which also runs strategy evolution (\cref{sec:evolve}) and regenerates trajectories for SFT (\cref{sec:sft}); it is never trained and is not used at evaluation or inference.
The harness logs every step: the reasoning, each tool call with its complete arguments (Write and Edit include full file contents or replaced strings), each observation (images included), and token usage.
Strategy internalization changes these data in two ways: a trajectory is replaced by a guided regeneration when the latter has a higher final score minus costs, and the playbook is added to the input (\cref{sec:sft}).

\Cref{tab:examples} summarizes three trajectories from the pool, one per task type.
Design images are tall (up to 5,562 pixels), so agents cut them into slices before reading them; 40.4\%, 42.3\%, and 5.8\% of the image observations in \cref{tab:examples} are design slices, screenshots of the agent's own pages, and contact sheets of the provided assets.

\begin{table}[tp]
  \centering
  \caption{Example trajectories from the pool, one per task type. --: not applicable; static tasks have no workflows.}
  \label{tab:examples}
  \small
  \setlength{\tabcolsep}{5pt}
  \begin{threeparttable}
  \begin{tabular}{@{}lccc@{}}
    \toprule
    & Static & Interactive & Full-stack \\
    \midrule
    \multicolumn{4}{@{}l}{\textit{Task}} \\
    Expected pages (behind sign-in) & 1 (3 viewports) & 6 & 11 (10) \\
    Workflows & -- & 7 & 13 \\
    Expected functional outcomes & -- & 50 & 105 \\
    \midrule
    \multicolumn{4}{@{}l}{\textit{Trajectory}} \\
    Assistant turns & 70 & 88 & 110 \\
    Tool calls & 89 & 94 & 145 \\
    Image observations & 24 & 28 & 52 \\
    Write / Edit calls & 6 / 9 & 22 / 0 & 10 / 26 \\
    Files edited & 6 & 22 & 10 \\
    Duration (min) & 60 & 39 & 82 \\
    \midrule
    \multicolumn{4}{@{}l}{\textit{Cost and outcome}} \\
    Input tokens & 5.3M & 9.9M & 20.4M \\
    Output tokens & 64K & 91K & 179K \\
    Filter verdict in the original run & pass & fail$^\dagger$ & pass \\
    \bottomrule
  \end{tabular}
  \begin{tablenotes}[flushleft]\footnotesize
    \item[$\dagger$] A browser driver was missing from the original run's environment; the trajectory passed when re-rendered.
  \end{tablenotes}
  \end{threeparttable}
\end{table}

\Cref{tab:pool} gives per-type means over passing trajectories.

\begin{table}[tp]
  \centering
  \caption{Means over the passing trajectories of the pool, by task type. --: not applicable.}
  \label{tab:pool}
  \small
  \setlength{\tabcolsep}{5pt}
  \begin{tabular}{lccc}
    \toprule
    & Static & Interactive & Full-stack \\
    \midrule
    \multicolumn{4}{l}{\textit{Task}} \\
    Workflows & -- & 7.63 & 12.96 \\
    Expected functional outcomes & -- & 53.88 & 63.50 \\
    \midrule
    \multicolumn{4}{l}{\textit{Trajectory}} \\
    Assistant turns & 70.48 & 82.06 & 99.55 \\
    Tool calls & 89.57 & 118.65 & 132.30 \\
    Image observations & 29.93 & 37.85 & 39.94 \\
    \bottomrule
  \end{tabular}
\end{table}

\section{Task Construction Pipeline}
\label{sec:supp_pipeline}

The candidate tasks of strategy evolution (\cref{sec:evolve}) are produced by a pipeline that (i) abstracts layout and interaction patterns from source websites and re-instantiates them in fictitious domains, (ii) writes the textual description, routes, and workflows and confirms that they describe the same application, (iii) has a teacher build a reference implementation, (iv) re-renders it from an isolated copy to produce the design images, and (v) admits the task only after engineering, visual, and functional quality control.
Asset file names are anonymized and visually similar distractor assets are added so that agents cannot infer image content from names.
Admitted tasks are divided into subsets for contrasting trajectory segments, small-batch testing, and validation, split by source website family to prevent near-duplicates among them; none shares a source website with an evaluation task.

\section{Checks and Checkers}
\label{sec:supp_checkers}

The checks of a task come from its files, and a checker $v_k(s)\in\{0,1\}$ decides each check $k$ after building and starting the website of $s$; a website that cannot be built or started fails every check, and $\Phi(s)$ is the fraction of checks passed (\cref{sec:formulation}).
\begin{itemize}[leftmargin=*,itemsep=1pt]
  \item \textbf{Functional checks (interactive and full-stack tasks).} Each workflow lists actions (\eg, ``Open /login'', ``Enter the seeded e-mail'', ``Click Sign in'') and expected outcomes (\eg, ``The browser leaves /login and lands on /''). Every expected outcome is one check: we replay the actions in a browser at the workflow's resolution, then judge the expected outcome with DOM queries when it names a route, element, or text, and with an MLLM on the relevant screenshot region otherwise. Workflows that depend on earlier ones (\eg, sign-in) are replayed in order.
  \item \textbf{Visual checks (all tasks).} Each page's design image, and for static tasks each of the desktop, tablet, and mobile images (1920, 1024, and 375 pixels wide), is split into its visual sections; each section is one check, judged by an MLLM on the region pair of design and rendered page.
  \item \textbf{Errors.} $\mathrm{Err}(s)$ counts console errors, failed resource loads, and failed builds or server starts. Errors are recorded at every evaluated state but do not enter $\Phi(s)$; a verification edit with $l_t>0$ and $\Delta\mathrm{Err}_t\le 0$ (\cref{eq:gainloss}) is a silent regression.
\end{itemize}

We apply the checkers at the end of coding ($s_{\mathrm C}$), after each verification edit (\cref{eq:verify}), and at the end of each re-run of \cref{eq:information}.
States before $s_{\mathrm C}$ are not rendered: perception steps are scored by re-runs (\cref{eq:information}) and coding steps by the evaluator (\cref{eq:coding}), with $\Phi(s_{\mathrm C})$ in place of $\hat\Phi_t$ for the last coding edit.

Each trajectory is replayed in an isolated copy of its initial workspace with a dedicated port.
Write and Edit calls are applied in the logged order; shell commands are re-executed in the sandbox, and any change they make to the website's files is recorded as a shell edit by diffing the workspace before and after. Trajectories whose commands depend on state outside the workspace are excluded, and the final replayed workspace is compared with the delivered workspace stored by the pipeline.
Before each render we wait until fonts, images, and lazy-loaded content have settled, and we store the hash of the source tree and of every screenshot so that each measurement can be reproduced.
We also fix the browser, the font files and font configuration, the viewports, and the waiting and scrolling policy: we found that a different font configuration alone changed page heights, which would otherwise register as a visual regression.
On the three example trajectories of \cref{tab:examples}, the reconstructed final workspaces match all 228 delivered application files byte for byte, and two independent captures of their 20 page--viewport combinations are pixel-identical.
Visual sections follow strong horizontal whitespace or background-color boundaries; sections shorter than 192 pixels are merged with the adjacent block, and sections taller than 768 pixels are split at the strongest internal boundary. \Cref{tab:checker_agreement} reports agreement for each checker type.

\begin{table}[tp]
  \centering
  \caption{Cohen's $\kappa$ between each checker and three human annotators on 200 samples per checker type.}
  \label{tab:checker_agreement}
  \small
  \setlength{\tabcolsep}{5pt}
  \begin{tabular}{lccc}
    \toprule
    Checker & A1 & A2 & A3 \\
    \midrule
    DOM functional & 0.97 & 0.96 & 0.98 \\
    Visual region & 0.78 & 0.76 & 0.81 \\
    Interaction outcome & 0.84 & 0.82 & 0.87 \\
    \bottomrule
  \end{tabular}
\end{table}

For the re-run reliability check, we sample 120 task--unit pairs across the three task types and compute each $\Delta_u$ independently from two disjoint sets of four re-runs; their Pearson correlation is 0.87.

\section{Pilot Study Details}
\label{sec:supp_pilot}

\subsection{Data and Measurement}

The teacher (\cref{sec:supp_agents}) works with the Read, Write, Edit, Bash, and task-list tools of its harness; its trajectories average about 70--100 assistant turns and 90--130 tool calls per task, varying by task type (\cref{tab:pool}).

\noindent\textbf{Hybrid outcome filter.}
We filter each delivered website with a two-stage DOM--GUI-agent pipeline.
The first stage builds and starts the website in an isolated environment and applies inexpensive deterministic gates derived from the task specification: the server must remain responsive, every public required route must return a nonempty document rather than an error page, and the DOM elements needed to begin each workflow must be present and actionable.
A failed build or server start, an uncaught page-level exception, or a failed gate immediately rejects the trajectory before any GUI-agent call.

Websites that pass the DOM gates proceed to functional replay.
A GUI agent observes the screenshot and accessibility tree and executes each specified workflow within an action budget of $\min(2L+4,30)$, where $L$ is the number of actions in the reference workflow.
After each action, deterministic outcomes---including the URL, element presence and visibility, text, form values, and persisted application state---are verified by DOM or backend queries; only outcomes that cannot be expressed reliably in this form are judged from the relevant screenshot region by the MLLM checker.
Independent workflows start from the same seeded state, whereas explicitly dependent workflows, such as sign-in followed by an authenticated operation, are replayed in order.
We retry a workflow once only when the browser or GUI agent reports an execution failure unrelated to the website; a checker failure is not retried.

An interactive or full-stack website passes the filter only if all DOM gates pass, every critical workflow reaches its final expected state, the functional and visual pass rates are at least 0.80 and 0.70, and every expected page is captured.
Critical workflows are the task-defining operations specified in the task files: authentication when required, the primary create or update action, and persistence of its result.
For static tasks, the functional conditions are omitted, while the build, route, page-capture, and visual pass-rate requirements remain.
To limit cost, workflows are evaluated with critical ones first, and evaluation stops as soon as a critical workflow fails or the best possible functional score after the remaining checks falls below 0.80.
This cascade reserves GUI-agent and MLLM calls for websites that are operational, while the hard critical-workflow gate prevents the aggregate threshold from accepting a visually plausible but unusable website.
The 7,284 passing and 2,416 failing trajectories form $\mathcal{D}^{+}$ and $\mathcal{D}^{-}$.

We reconstruct each $s_t$ by replaying the logged edits on an isolated copy of $s_0$ (\cref{sec:supp_checkers}): Write and Edit calls replay exactly from their recorded file contents or replaced strings, and shell commands are re-executed in the sandbox.
Trajectories whose replayed final workspace differs from the stored one are discarded (2.4\%).
We evaluate the website only at $s_{\mathrm C}$ and after each edit step in verification (a verification edit): an MLLM judges visual checks on pairs of design and page regions, and DOM queries or an MLLM judge functional checks.
We record the check results $v_k(s)$, which give $\Phi(s)$, and the number $\mathrm{Err}(s)$ of console errors, failed resource loads, and failed builds or server starts.
Since only edit steps change the website, $s_{t-1}$ equals the previous evaluated state for every verification edit $t$; the fractions of checks it newly passes and breaks are
\begin{equation}
  \begin{aligned}
    g_t&=\frac{1}{K}\sum_{k=1}^{K}v_k(s_t)\bigl[1-v_k(s_{t-1})\bigr],\\
    l_t&=\frac{1}{K}\sum_{k=1}^{K}v_k(s_{t-1})\bigl[1-v_k(s_t)\bigr],
  \end{aligned}
  \label{eq:gainloss}
\end{equation}
so that $b_t=g_t-l_t$ (\cref{eq:verify}); other verification steps have $g_t=l_t=0$.
An edit with $l_t>0$ causes a regression; a verification edit is productive if it passes a new check without a regression or more errors, $e_t=\ind[g_t>0,\ l_t=0,\ \Delta\mathrm{Err}_t\le 0]$ with $\Delta\mathrm{Err}_t=\mathrm{Err}(s_t)-\mathrm{Err}(s_{t-1})$.

\subsection{What the Outcome Misses}

Summing \cref{eq:gainloss} over the verification steps $\mathcal{V}$ telescopes over the evaluated states:
\begin{equation}
  \Phi(s_T)-\Phi(s_{\mathrm C})=\sum_{t\in\mathcal{V}}b_t=\sum_{t\in\mathcal{V}}g_t-\underbrace{\sum_{t\in\mathcal{V}}l_t}_{W(\tau)} .
  \label{eq:telescope}
\end{equation}
Regressions accumulate, with repeats, into the rework $W(\tau)$; since $\sum_{t\in\mathcal{V}}g_t=\Phi(s_T)-\Phi(s_{\mathrm C})+W(\tau)$, every unit of rework requires an equal extra gain.
Stagnant steps, with $g_t=l_t=0$, leave no trace in \cref{eq:telescope}.
The final score or the outcome filter's verdict, all that outcome-level supervision sees, thus admits different amounts of rework and stagnation, the more so the longer the horizon.

The execution efficiency $\eta(\tau)$ is the fraction of productive edits among the verification edits of $\tau$, excluding trajectories without such edits (3.7\%).
Within a horizon bin (a task type, or a tercile of $T$ within a type), with $\tau^{+}$ and $\tau^{-}$ drawn from its passing and failing trajectories $\mathcal{D}^{+}$ and $\mathcal{D}^{-}$,
\begin{equation}
  \mathrm{AUC}=\Pr\big[\eta(\tau^{+})>\eta(\tau^{-})\big]+\tfrac{1}{2}\Pr\big[\eta(\tau^{+})=\eta(\tau^{-})\big]
  \label{eq:auc}
\end{equation}
measures how well process quality orders outcomes, with chance level $0.5$, and the \emph{lucky-success rate}
\begin{equation}
  \mathrm{LS}=\frac{\big|\{\tau\in\mathcal{D}^{+} : \eta(\tau)<\operatorname{med}_{\tau'\in\mathcal{D}^{-}}\eta(\tau')\}\big|}{|\mathcal{D}^{+}|}
  \label{eq:ls}
\end{equation}
is the fraction of successes with process quality below the failure median.
\Cref{tab:pilot} reports both by task type.

\begin{table}[tp]
  \centering
  \caption{Process statistics of teacher trajectories by task type. Rows use the trajectories that pass the outcome filter, except the pass rate, lucky-success rate (\cref{eq:ls}), and AUC (\cref{eq:auc}).}
  \label{tab:pilot}
  \setlength{\tabcolsep}{5pt}
  \small
  \begin{tabular}{lccc}
    \toprule
    & Static & Interactive & Full-stack \\
    \midrule
    Steps $T$ & 70.5 & 82.1 & 99.6 \\
    Pass rate of the outcome filter & 81.4 & 74.8 & 65.7 \\
    Execution efficiency $\eta$ & 0.61 & 0.48 & 0.36 \\
    Rework $W(\tau)$ & 0.031 & 0.057 & 0.094 \\
    Verification edits with $l_t>0$ (\%) & 7.8 & 12.6 & 18.9 \\
    \quad silent among them ($\Delta\mathrm{Err}_t\le 0$, \%) & 71.4 & 78.2 & 84.7 \\
    Lucky-success rate LS (\%) & 8.6 & 19.7 & 34.1 \\
    AUC & 0.84 & 0.69 & 0.55 \\
    \bottomrule
  \end{tabular}
\end{table}

\subsection{Further Results}

Within each task type, AUC also falls from the shortest to the longest tercile of $T$: from 0.89 to 0.77 on static tasks, 0.75 to 0.61 on interactive tasks, and 0.62 to 0.50 on full-stack tasks (\cref{find:decoupling}); in passing full-stack trajectories 57.3\% of verification edits pass no new check ($g_t=0$).

\begin{figure*}[tp]
  \centering
  \includegraphics[width=0.95\linewidth]{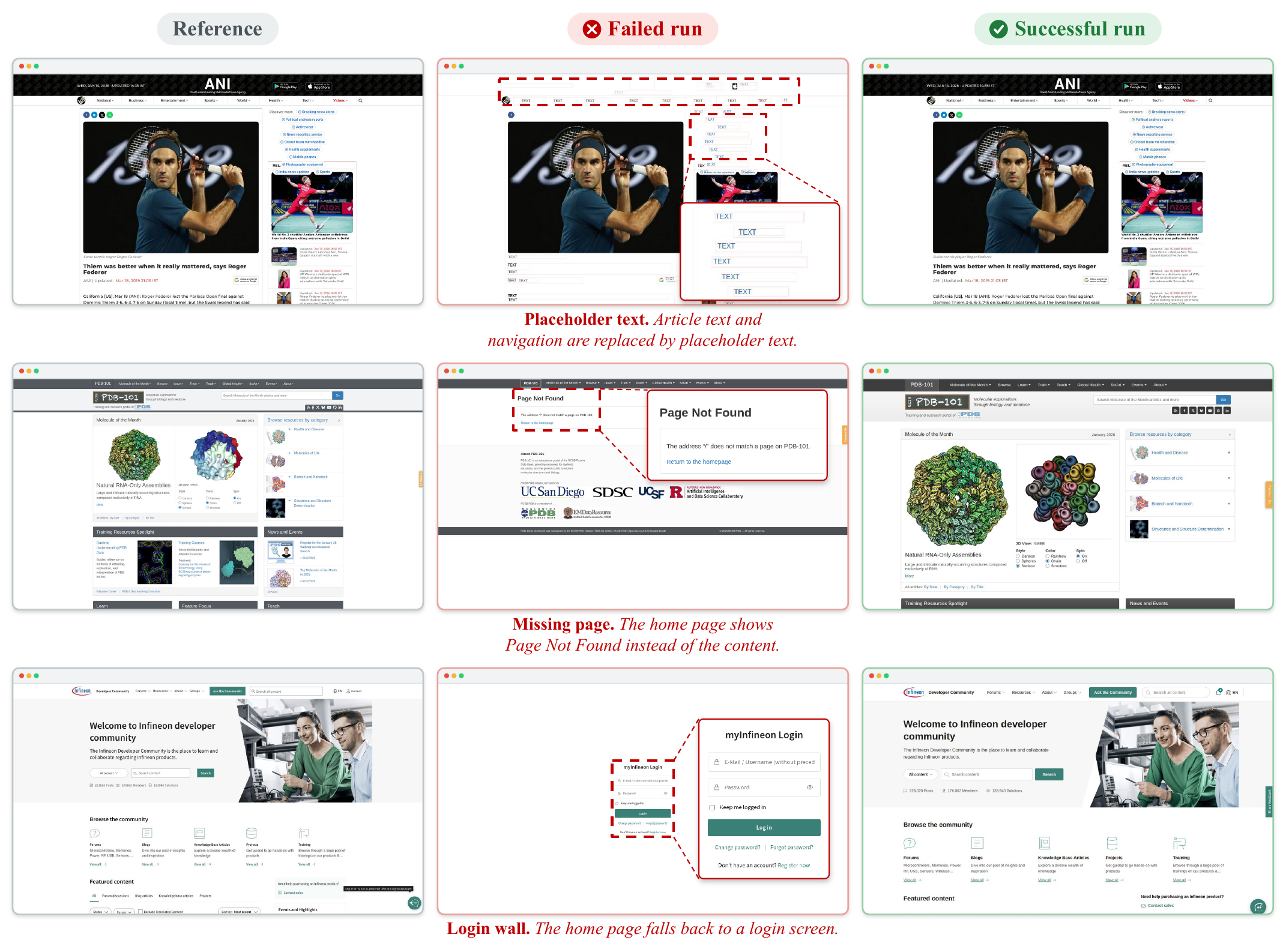}
  \caption{\textbf{Typical failures in visual coding.} Each row shows one task: the design, a run whose delivered page fails visibly (red box, magnified in the inset), and a run that reproduces the design.}
  \label{fig:qualitative}
\end{figure*}

\Cref{fig:qualitative} shows typical failures that reach the delivered website when the agent does not look at its page after the last edit; a single screenshot would reveal each of them.

\section{Behavior Descriptors}
\label{sec:supp_behaviors}

\Cref{tab:behaviors} lists the rules that derive the behavior types of each step from the action log; the edit types define the operations whose step rewards \cref{find:frequency} compares across stages.

\begin{table}[tp]
  \centering
  \caption{Behavior descriptors extracted from action logs.}
  \label{tab:behaviors}
  \small
  \setlength{\tabcolsep}{5pt}
  \begin{tabular}{p{0.3\linewidth}p{0.64\linewidth}}
    \toprule
    Descriptor & Rule \\
    \midrule
    \multicolumn{2}{l}{\textit{Edits}} \\
    Edit type & Create a file (Write to a new path), rewrite a file (Write to an existing path), local edit (Edit), shell edit (a Bash command that changes the website's files), or asset edit (an asset cropped from the design). \\
    Evidence freshness & Stale if the edited file changed after the agent last read it; mismatch errors from the editing tool are recorded separately. \\
    Repetition & The edit targets the same element and applies the same pattern as an earlier edit whose contribution was not positive (\cref{eq:coding,eq:verify}). \\
    \midrule
    \multicolumn{2}{l}{\textit{Observations}} \\
    Observation type & Design slice, asset contact sheet, own screenshot, side-by-side comparison image, zoomed crop, or command output (build, server start, or HTTP request). Images are identified by pixel hash rather than file name, since agents regenerate comparison images under the same name. \\
    \midrule
    \multicolumn{2}{l}{\textit{Completion}} \\
    Completion claim & A task-list item is marked completed (TaskUpdate), or the agent declares the task done in its final message. \\
    \bottomrule
  \end{tabular}
\end{table}

\section{Coding Evaluator}
\label{sec:supp_evaluator}

The evaluator that produces $\hat v_k$ in \cref{eq:coding} is Kimi-K2.7-Code~\cite{moonshot2026kimik27code}, used as a fixed MLLM judge whose parameters are not updated during policy training.
We use deterministic decoding and require a JSON response.
The evaluator is called after each edit in the coding stage (\cref{sec:formulation}), and other coding steps receive $b_t=0$.
From the end of coding onward, contributions are computed directly by \cref{eq:verify}.

\noindent\textbf{Input organization.}
For a coding step $t$, we construct the evaluator input from four ordered blocks:
\begin{enumerate}[leftmargin=*,itemsep=1pt]
  \item \textsc{Task}: the user request and the reference design images;
  \item \textsc{Requirements}: the same $K$ atomic visual and functional requirements used by the task checkers, each with a stable identifier, its type, and its success condition;
  \item \textsc{History}: the textual and visual observations available through step $t$, in chronological order, with each observation labeled by its step and tool source; repeated identical images are deduplicated by pixel hash; and\looseness=-1
  \item \textsc{Workspace}: a lexicographically sorted file manifest followed by the full contents of the website's source and configuration files, with each text file enclosed in a path- and language-labeled block. Dependency directories, build outputs, caches, and other generated files are omitted. Binary assets are represented by path, media type, and dimensions, while assets already viewed by the agent remain available as images in \textsc{Task} or \textsc{History}.
\end{enumerate}
The input contains no future actions, final task score, or checker outcomes unavailable at step $t$.
Thus, the evaluator uses no evidence from after step $t$; beyond the policy's context, it receives only the requirement list and an explicit view of the current workspace $s_t$.

\noindent\textbf{Evaluation prompt.}
The following instruction is prepended to every evaluator request:
\begin{quote}\small
You are a coding-progress evaluator.
Given a web-development task, its atomic requirements, the observations available so far, and the current workspace, assess how completely the current code implements each requirement.
Judge the implementation state rather than the quality of the agent's plan or explanation.
Do not award credit for intentions, task-list entries, comments, unused components, or code that is not connected to the delivered application.
Use visual and textual observations only as evidence available at the current step, and do not assume that an unobserved behavior works.

Score every requirement independently in $[0,1]$ using the following anchors: 0 means absent or contradicted; 0.25 means only scaffolding or a small relevant fragment is present; 0.5 means the main mechanism is present but important parts are missing; 0.75 means the requirement is mostly implemented with minor gaps; and 1 means it is fully implemented and consistent with all available evidence.
Intermediate values are allowed.
For each requirement, cite concise evidence from file paths, code elements, or observations, and state the main missing part when the score is below 1.
Return only JSON with one entry for every supplied requirement, preserving the supplied identifiers and order.
\end{quote}

The required output schema is a top-level \texttt{requirements} list whose entries contain \texttt{id}, \texttt{score}, \texttt{evidence}, and \texttt{missing} fields.
We accept an evaluation only when it contains each of the $K$ identifiers exactly once and every score is a finite value in $[0,1]$.
If validation fails, we allow at most two retries with a schema-correction message; if neither retry produces a valid response, the contribution of that edit is set to $b_t=0$.
Only the scores enter $\hat\Phi_t$ in \cref{eq:coding}; evidence and missing-part fields are retained for auditing.

\section{An Illustrative Trajectory}
\label{sec:supp_case}

\begin{figure*}[tp]
  \centering
  \includegraphics[width=\linewidth]{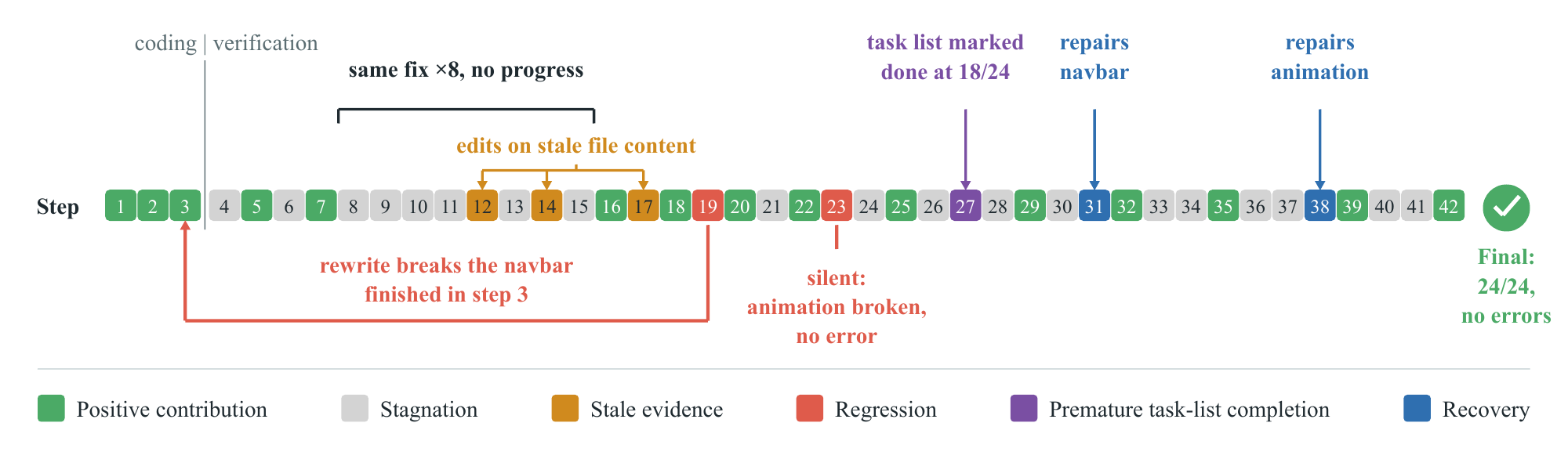}
  \caption{The 42 coding and verification steps of a successful trajectory on a static product landing page with 24 checks in 6 files. Green steps have positive contributions $b_t$ and red regressions negative ones, which the recoveries (blue) cancel; stagnant steps (gray), edits on stale evidence (amber), and the premature task-list completion (purple) contribute nothing.}
  \label{fig:case42}
\end{figure*}

\Cref{fig:case42} shows the 42 coding and verification steps of a successful trajectory (its perception steps, credited by \cref{eq:information}, are omitted), and \cref{tab:case42} describes its annotated events with the contributions that step-level evaluation assigns to them (\cref{sec:attr}).
Coding ends after step 3, and from step 4 on the agent renders its page: each verification edit is credited with the change in score (\cref{eq:verify}), and steps that only inspect the page receive $b_t=0$.

\begin{table}[tp]
  \centering
  \caption{Annotated events of \cref{fig:case42} and their contributions.}
  \label{tab:case42}
  \small
  \setlength{\tabcolsep}{5pt}
  \begin{tabular}{>{\raggedright\arraybackslash}p{0.17\linewidth}>{\raggedright\arraybackslash}p{0.52\linewidth}>{\raggedright\arraybackslash}p{0.23\linewidth}}
    \toprule
    Steps & Event & Contribution \\
    \midrule
    \multicolumn{3}{l}{\textit{Stagnant steps}} \\
    8--15 & The form layout is misaligned; the same fix is attempted eight times without changing which checks pass. & $b_t=0$, $r_t<0$ \\
    12, 14, 17 & Replacements based on file content read about ten steps earlier, before later edits changed it, fail; the files do not change. & $b_t=0$, $r_t<0$ \\
    \midrule
    \multicolumn{3}{l}{\textit{Regressions and repairs}} \\
    19 & Rewriting \texttt{styles.css} during verification breaks the navigation bar finished in step 3. & $b_{19}<0$ \\
    23 & Restructuring the hero section renames a class used by an animation script; the animation stops without any error. & $b_{23}<0$ \\
    31, 38 & The navigation bar and the animation are repaired. & $b_{31}=-b_{19}$, $b_{38}=-b_{23}$ \\
    \midrule
    \multicolumn{3}{l}{\textit{Premature completion}} \\
    27 & The agent marks its last task-list item completed with 18 of 24 checks passed but continues. & $b_{27}=0$ \\
    \bottomrule
  \end{tabular}
\end{table}

Trajectory-level supervision credits all 42 steps alike, since the final page passes all 24 checks without errors (\cref{fig:teaser}, top).
Step-level evaluation separates them.
The stagnant steps, including the failed edits on stale file content, contribute nothing but spend a turn and tokens, so their step rewards are negative (\cref{eq:utility}).
The regressions at steps 19 and 23 receive negative contributions, and the repairs at steps 31 and 38 receive positive contributions of the same size; each pair cancels in the sum of the verification contributions (\cref{eq:verify}), while all four steps pay for their turns and tokens.
Step 23 leaves $\mathrm{Err}(s)$ unchanged (\cref{sec:supp_checkers}), so error signals miss the regression, while its negative contribution exposes it.
The task-list update at step 27 changes no website file and receives $b_{27}=0$; had the agent stopped there, the final score would have been $18/24$.

\section{Evolved Playbook and Adoption}
\label{sec:supp_guidance}

Strategy evolution (\cref{sec:evolve}) starts from an empty $G$ and proposes one bounded revision per iteration.
\Cref{tab:revisions} lists each revision, the lower confidence bound of its improvement on the validation tasks, and its outcome under \cref{eq:deficit}.
\Cref{tab:guidance} lists the rules of the final playbook, which strategy internalization uses (\cref{sec:inject}).

\begin{table}[tp]
  \centering
  \caption{Revisions during strategy evolution. Rules: numbers of rules added, edited, and removed; LCB: 95\% lower confidence bound of $J(G')-J(G)$ on the validation tasks; decision: accepted, rejected and recorded, or discarded before evaluation for repeating a recorded rejection; $J(G)$: final score minus costs on the validation tasks with the playbook after the iteration (\cref{eq:validation}).}
  \label{tab:revisions}
  \small
  \begin{tabular}{ccccc}
    \toprule
    Iteration & Rules & LCB & Decision & $J(G)$ \\
    \midrule
    0 & -- & -- & initialization & 0.701 \\
    1 & 2/0/0 & $+0.006$ & accepted & 0.711 \\
    2 & 1/1/0 & $+0.008$ & accepted & 0.722 \\
    3 & 0/1/0 & $-0.003$ & rejected & 0.722 \\
    4 & 1/0/0 & $+0.006$ & accepted & 0.730 \\
    5 & 1/1/0 & $+0.004$ & accepted & 0.738 \\
    6 & 0/1/0 & $+0.001$ & rejected & 0.738 \\
    7 & 1/1/0 & $+0.005$ & accepted & 0.746 \\
    8 & 0/1/0 & -- & discarded & 0.746 \\
    9 & 0/2/0 & $+0.003$ & accepted & 0.752 \\
    10 & 0/1/0 & $-0.001$ & rejected & 0.752 \\
    11 & 0/1/0 & -- & discarded & 0.752 \\
    12 & 0/1/0 & $+0.001$ & rejected & 0.752 \\
    \bottomrule
  \end{tabular}
\end{table}

\begin{table}[tp]
  \centering
  \caption{Rules of the final playbook in if--then form, grouped by the part of the task they guide.}
  \label{tab:guidance}
  \small
  \setlength{\tabcolsep}{5pt}
  \begin{tabular}{l>{\raggedright\arraybackslash}p{0.42\linewidth}>{\raggedright\arraybackslash}p{0.46\linewidth}}
    \toprule
    \# & If & then \\
    \midrule
    \multicolumn{3}{l}{\textit{Reviewing the design and the assets}} \\
    1 & no code has been written yet, & read every design image individually at full resolution. \\
    2 & several inventories, image-size scans, or measurements are independent, & combine them into batched checks. \\
    3 & the provided assets are numerous, & review them in labeled contact sheets, without reopening images to recheck details already observed. \\
    \midrule
    \multicolumn{3}{l}{\textit{Writing the code}} \\
    4 & the design and the assets have been reviewed, & start writing the page structure and styles, estimating minor visual details and refining them later in the running page. \\
    5 & a required asset is missing, & crop it from a design image, save it as an asset file, and reference that file; otherwise reference the original asset paths.\looseness=-1 \\
    \midrule
    \multicolumn{3}{l}{\textit{Verifying the page}} \\
    6 & the code has been written and deployed, & screenshot the page, compare it with the design, fix the discrepancies, and stop after a few focused cycles once the page works and the main discrepancies are resolved.\looseness=-1 \\
    \bottomrule
  \end{tabular}
\end{table}

\Cref{tab:adoption} reports the adoption rates of rules 3 and 5, whose behaviors \cref{find:adoption} shows to be rare in successful trajectories.
The adoption rate of a rule is the fraction of trajectories in which the agent carries out the rule's behavior while its condition holds, among the trajectories in which the condition holds at all; the joint rate is the fraction of trajectories carrying out both behaviors among those in which both conditions hold.
The table compares our base model (\cref{sec:pilot}), without and with the playbook in its context, with \method (no playbook at inference).

We measure adoption on all 193 evaluation tasks, with deterministic rules on the action log that detect the behaviors of the playbook.
Behaviors are located relative to the first application edit, which ends perception (\cref{sec:formulation}); it is detected as the first successful write or edit of a page source file (HTML, CSS, JavaScript or TypeScript, or a framework component), including writes by shell commands, but not the agent's own helper scripts or inspection images.
Operationally, the conditions of rules 1 and 2 hold in every trajectory before its first application edit and that of rule 6 after it; the condition of rule 3 holds when the task provides at least 8 assets, and that of rule 5 as defined below.
We record whether each behavior occurs in a trajectory and count screenshot--edit--screenshot cycles:
\begin{itemize}[leftmargin=*,itemsep=1pt]
  \item \textbf{Complete design inspection}: before the first application edit, we record whole-image reads of every design image (rule 1) and, separately, views of design slices.
  \item \textbf{Batched inventory} (rule 2): a single shell command before the first application edit performs at least two inventory operations (listing, searching, or measuring files) on the design images or assets.
  \item \textbf{Image-size scan} (rule 2): a shell command before the first application edit reads the sizes of design images or provided assets.
  \item \textbf{Contact sheet} (rule 3): a command composes several images into one with text labels, and the agent views that output, both before the first application edit.
  \item \textbf{Asset cropped from the design} (rule 5): a design image is cropped, the crop is saved as an asset file, and an application file references it; the condition of rule 5 holds when the maximum CLIP cosine similarity between the region and any provided asset, after aspect-ratio-preserving resize, is below 0.82.
  \item \textbf{Page screenshot} (rule 6): after the first application edit, a screenshot command followed by an image read that is not a design image, asset, or contact sheet.
  \item \textbf{Screenshot--edit--screenshot cycle} (rule 6): two consecutive such pairs with an application edit between the first read and the second screenshot.
  \item \textbf{Final look} (rule 6): a page image read after the last application edit.
  \item \textbf{Full strategy bundle}: every design image is read before coding, a batched inventory and a viewed labeled contact sheet are used, and at least one screenshot--edit--screenshot cycle is completed in the same trajectory.
\end{itemize}
Rule 4 has no reliable detector and is not reported.
\Cref{tab:behavior} gives the rates shown in \cref{fig:behavior}, together with those of \method with $G$ appended to its prompt, and \cref{tab:behavior_levels} breaks them down by task type. The base model is a separate evaluation run; four of its full-stack logs are incomplete and excluded, and its detector additionally recognizes images created by shell scripts.
\method with $G$ appended to its prompt at inference is another separate run, analyzed with the same detector as the base model; for whole-image reads, 1, 2, and 1 of its trajectories per task type cannot be determined and are excluded.

\begin{table}[tp]
  \centering
  \caption{Behavior of the agent through training. Share of the evaluation trajectories that show each behavior (\%, unless noted), for our base model, \method after its SFT stage with the playbook $G$ as privileged input (SFT), and \method after RL, all without $G$ at inference; the last column appends $G$ to the prompt of \method after RL. The base model is measured on the 189 of 193 trajectories with complete logs. Screenshot cycle: screenshot, edit, screenshot. Rule numbers refer to \cref{tab:guidance}.}
  \label{tab:behavior}
  \small
  \setlength{\tabcolsep}{5pt}
  \begin{tabular}{lc>{\columncolor{ourrow}}c>{\columncolor{ourrow}}c>{\columncolor{ourrow}}c}
    \toprule
    & & \multicolumn{3}{c}{\method} \\
    \cmidrule(lr){3-5}
    Behavior (rule) & Base & \multicolumn{1}{c}{SFT} & \multicolumn{1}{c}{+ RL} & \multicolumn{1}{c}{+ $G$} \\
    \midrule
    \multicolumn{5}{l}{\textit{Before coding}} \\
    Complete design inspection (1) & 97.9 & 25.4 & 87.0 & 43.9 \\
    Batched inventory (2) & 68.8 & 100.0 & 92.7 & 100.0 \\
    Image-size scan (2) & 13.8 & 100.0 & 84.5 & 100.0 \\
    Contact sheet (3) & 1.1 & 49.2 & 34.7 & 62.7 \\
    Design viewed in slices & -- & 79.8 & 48.2 & 76.2 \\
    \midrule
    \multicolumn{5}{l}{\textit{After coding starts}} \\
    Page screenshot (6) & 15.3 & 98.4 & 89.1 & 100.0 \\
    $\ge$1 screenshot cycle (6) & 13.2 & 94.3 & 82.9 & 88.6 \\
    Final look (6) & 10.6 & 71.0 & 68.4 & 89.1 \\
    Full strategy bundle & 1.1 & 9.3 & 23.8 & -- \\
    \midrule
    \multicolumn{5}{l}{\textit{Iteration and effort}} \\
    $\ge$4 screenshot cycles & 2.1 & 67.9 & 34.7 & 49.2 \\
    Tool calls per task (mean) & 101.3 & 267.1 & 159.4 & 207.9 \\
    Tool calls per task (median) & 89 & 196 & 140 & 171 \\
    \bottomrule
  \end{tabular}
\end{table}

\begin{table*}[tp]
  \centering
  \caption{Behavior of the models of \cref{tab:behavior} by task type (\% of trajectories unless noted). S, I, and F: static webpages, interactive websites, and full-stack websites, with 100, 66, and 27 tasks (23 full-stack tasks with complete logs for the base model); shaded columns are \method-SFT, \method, and \method with $G$ in its prompt at inference (${}+G$); rule numbers refer to \cref{tab:guidance}.}
  \label{tab:behavior_levels}
  \small
  \setlength{\tabcolsep}{2.4pt}
  \begin{tabular}{lccc>{\columncolor{ourrow}}c>{\columncolor{ourrow}}c>{\columncolor{ourrow}}c>{\columncolor{ourrow}}c>{\columncolor{ourrow}}c>{\columncolor{ourrow}}c>{\columncolor{ourrow}}c>{\columncolor{ourrow}}c>{\columncolor{ourrow}}c}
    \toprule
    & \multicolumn{3}{c}{Base model} & \multicolumn{3}{c}{\method-SFT} & \multicolumn{3}{c}{\method} & \multicolumn{3}{c}{\method${}+G$} \\
    \cmidrule(lr){2-4}\cmidrule(lr){5-7}\cmidrule(lr){8-10}\cmidrule(lr){11-13}
    Behavior (rule) & S & I & F & \multicolumn{1}{c}{S} & \multicolumn{1}{c}{I} & \multicolumn{1}{c}{F} & \multicolumn{1}{c}{S} & \multicolumn{1}{c}{I} & \multicolumn{1}{c}{F} & \multicolumn{1}{c}{S} & \multicolumn{1}{c}{I} & \multicolumn{1}{c}{F} \\
    \midrule
    \multicolumn{13}{l}{\textit{Before the first application edit}} \\
    Batched inventory (2) & 56.0 & 84.8 & 78.3 & 100.0 & 100.0 & 100.0 & 87.0 & 98.5 & 100.0 & 100.0 & 100.0 & 100.0 \\
    Image-size scan (2) & 22.0 & 6.1 & 0.0 & 100.0 & 100.0 & 100.0 & 80.0 & 95.5 & 74.1 & 100.0 & 100.0 & 100.0 \\
    Contact sheet (3) & 2.0 & 0.0 & 0.0 & 40.0 & 53.0 & 74.1 & 22.0 & 51.5 & 40.7 & 64.0 & 63.6 & 55.6 \\
    Design sliced, slices viewed & -- & -- & -- & 80.0 & 77.3 & 85.2 & 48.0 & 51.5 & 40.7 & 72.0 & 83.3 & 74.1 \\
    Complete design inspection (1) & 97.0 & 98.5 & 100.0 & 39.0 & 4.5 & 25.9 & 96.0 & 75.8 & 81.5 & 60.6 & 23.4 & 30.8 \\
    \midrule
    \multicolumn{13}{l}{\textit{After the first application edit}} \\
    Page screenshot (6) & 19.0 & 13.6 & 4.3 & 99.0 & 98.5 & 96.3 & 82.0 & 97.0 & 96.3 & 100.0 & 100.0 & 100.0 \\
    $\ge$1 screenshot--edit--screenshot cycle (6) & 17.0 & 12.1 & 0.0 & 96.0 & 92.4 & 92.6 & 76.0 & 87.9 & 96.3 & 83.0 & 98.5 & 85.2 \\
    Final look (6) & 14.0 & 9.1 & 0.0 & 73.0 & 68.2 & 70.4 & 61.0 & 74.2 & 81.5 & 88.0 & 90.9 & 88.9 \\
    Full strategy bundle & 2.0 & 0.0 & 0.0 & 11.0 & 3.0 & 18.5 & 15.0 & 33.3 & 33.3 & -- & -- & -- \\
    \midrule
    \multicolumn{13}{l}{\textit{Iteration and effort}} \\
    $\ge$4 screenshot--edit--screenshot cycles & 3.0 & 1.5 & 0.0 & 64.0 & 71.2 & 74.1 & 24.0 & 39.4 & 63.0 & 38.0 & 57.6 & 70.4 \\
    Tool calls before coding (median) & 26 & 36 & 35 & 44 & 55 & 68 & 45 & 49 & 49 & 47 & 58 & 63 \\
    Tool calls per task (mean) & 69.7 & 120.4 & 183.7 & 162.2 & 262.9 & 666.2 & 102.1 & 179.0 & 324.1 & 136.5 & 221.8 & 438.6 \\
    Tool calls per task (median) & 58.5 & 104 & 171 & 140.5 & 244.5 & 585 & 92 & 157.5 & 302 & 118 & 196 & 401 \\
    \bottomrule
  \end{tabular}
\end{table*}

\begin{table}[tp]
  \centering
  \caption{Adoption rates (\%) of rules 3 and 5; rule numbers refer to \cref{tab:guidance}. Base: our base model, without the playbook; Base${}+G$: the same model with the playbook in its context; \method: no playbook at inference.}
  \label{tab:adoption}
  \small
  \setlength{\tabcolsep}{5pt}
  \begin{tabular}{lcc>{\columncolor{ourrow}}c}
    \toprule
    Rule & Base & Base${}+G$ & \multicolumn{1}{c}{\method} \\
    \midrule
    \multicolumn{4}{l}{\textit{Per rule}} \\
    3: labeled contact sheets & 2.4 & 72.4 & 44.8 \\
    5: crop a missing asset & 7.9 & 50.6 & 35.1 \\
    \midrule
    \multicolumn{4}{l}{\textit{Joint adoption}} \\
    Rules 3 and 5 & 4.8 & 63.9 & 42.6 \\
    \bottomrule
  \end{tabular}
\end{table}

\section{Hyperparameters}
\label{sec:supp_hparams}

Step-level evaluation uses $Q=4$ re-runs with and without each mask in \cref{eq:information}, $\lambda_{\mathrm{turn}}=1{\times}10^{-5}$ and $\lambda_{\mathrm{tok}}=1{\times}10^{-8}$ in \cref{eq:utility}.
Strategy evolution changes at most $m=2$ rules per revision and accepts a revision if the lower confidence bound of the paired bootstrap, at a 95\% confidence level with 2,000 resamples, exceeds $\gamma=0.002$ (\cref{eq:deficit}); it runs at most 12 iterations and stops after 3 consecutive iterations without an accepted revision.
All SFT variants run 400 optimizer steps with a global batch size of 64 and sequences packed to 256K tokens, using the Muon optimizer with a peak learning rate of $5{\times}10^{-5}$, 10 warm-up steps, cosine decay, and a 0.1$\times$ learning rate for the vision encoder and projection layers.
RL uses 200 tasks selected from the SFT data, $M=8$ rollouts per task, $\epsilon=10^{-6}$ in \cref{eq:rl_reward}, and AdamW with a peak learning rate of $2{\times}10^{-5}$, a global batch size of 16, a maximum context length of 256K tokens, and 25 optimizer steps.
Varying $\lambda_{\mathrm{turn}}$ over $\{5{\times}10^{-6},1{\times}10^{-5},2{\times}10^{-5}\}$ gives overall scores of 77.6, 78.2, and 77.4; varying $\lambda_{\mathrm{tok}}$ over $\{5{\times}10^{-9},1{\times}10^{-8},2{\times}10^{-8}\}$ gives 77.8, 78.2, and 77.5. Using $Q\in\{2,4,8\}$ gives 77.1, 78.2, and 78.3, so we use $Q=4$.

\section{Evaluation Details}
\label{sec:supp_eval}

Fine-tuning a pre-trained model on a new domain can erode abilities it already has, as observed in class-incremental learning~\cite{sun2025mos,sun2025pilot} and in the multilingual abilities of MLLMs after visual instruction tuning~\cite{sun2025parrot}, so we also evaluate \method on general multimodal benchmarks.
\Cref{tab:general} lists the scores shown in \cref{fig:general}. All models are evaluated with reasoning enabled.
Where a model's official report gives a score, we use it in place of our own evaluation.
Design2Code uses its official test split and mean GPT score; AndroidWorld uses the test split and task success rate; OmniDocBench v1.5 uses the official test set and the overall end-to-end score; CharXiv uses accuracy on the reasoning questions of its test set; MMMU-Pro uses the 10-option test split; MathVista uses testmini accuracy; and HLE uses exact-match accuracy on its multimodal subset.

\section{Extended Related Work}
\label{sec:supp_related}

This section extends the three topics of \cref{sec:related}, treating outcome-level training and dense credit separately, and adds a topic on verification and visual tool use.

\noindent\textbf{Visual coding.}
Generating code from visual inputs started with screenshot-to-markup translation~\cite{beltramelli2018pix2code} and grew into large datasets and benchmarks for webpage generation~\cite{laurencon2024websight,yun2024web2code,si2025design2code,gui2025webcode2m}.
Later benchmarks add interaction~\cite{xiao2025interaction2code}, functional websites built from scratch~\cite{lu2025webgenbench}, long pages~\cite{longwebbench2026}, ambiguous user instructions~\cite{interactweb2026}, and hierarchical tasks up to full-stack development with workflow-based agent verification~\cite{he2026vision2web}.
Beyond fine-tuning on image--code pairs~\cite{jiang2025viscodex}, training methods for visual code reward the rendered output: ScreenCoder optimizes the pixel similarity between the rendered page and the screenshot~\cite{jiang2025screencoder}, RRVF rewards visual similarity in a loop of reasoning, rendering, and visual feedback~\cite{chen2025rrvf}, other methods refine the code over rounds of visual feedback~\cite{ui2coden2025,visrefiner2026}, AesCoder rewards executability and static and interactive aesthetics through a multi-agent reward~\cite{xiao2025aescoder}, and AesCode, which generates slides, posters, and dashboards as code, scores parsable properties with deterministic verifiers and the rest with a VLM rubric~\cite{gao2026aescode}.
For websites, WebGen-Agent scores screenshots and GUI-agent tests at every step and uses these scores as step rewards in Step-GRPO~\cite{lu2026webgenagent}; ReLook trains with an MLLM critic and accepts only improving revisions~\cite{relook2025}; RILA replays reference interactions to obtain execution-aware observations~\cite{rila2026}; WebGen-R1 combines structural, functional, and aesthetic rewards~\cite{webgenr12026}.
All of these rewards score the output, and a step reward computed on the current page presumes that the website runs, which need not hold before coding ends.
\method therefore evaluates the reconstructed website from the end of coding on, scoring zero when it cannot be built or started (\cref{sec:supp_checkers}), credits coding steps with the estimated change in requirement implementation (\cref{eq:coding}) and perception steps with their information value (\cref{eq:information}), and charges every step for its turn and tokens (\cref{eq:utility}).\looseness=-1

\begin{table}[tp]
  \centering
  \caption{Transfer and general capability of \method, its base model, and open-source models. $^{*}$: official report; other scores are evaluated by us. Best in bold, second underlined.}
  \label{tab:general}
  \small
  \setlength{\tabcolsep}{5pt}
  \begin{tabular}{lccccccc}
    \toprule
    & \multicolumn{2}{c}{Coding \& agents} & \multicolumn{2}{c}{Documents} & \multicolumn{3}{c}{Reasoning} \\
    \cmidrule(lr){2-3}\cmidrule(lr){4-5}\cmidrule(l){6-8}
    Model & D2C & AW & ODB & CXR & MMMU & MV & HLE \\
    \midrule
    Qwen3.5-122B-A10B & 92.6 & 66.4 & 89.8 & 77.2 & 76.9 & 87.4 & 18.1 \\
    DeepSeek-V4.1-Flash & \secondscore{93.6} & \secondscore{73.3} & 79.9 & 71.1 & 75.6 & 85.2 & 20.8 \\
    GLM-5.3-Flash & 84.1 & 46.1 & 90.5 & 80.5 & 77.8 & 89.3 & \bestscore{28.1} \\
    Kimi-K3 & 92.6 & 52.6 & 89.9 & \bestscore{84.8}$^{*}$ & \bestscore{81.6}$^{*}$ & 89.4 & \secondscore{27.8} \\
    \midrule
    Our base model & 93.5 & 67.2 & \secondscore{91.4}$^{*}$ & 81.3$^{*}$ & 79.0$^{*}$ & \bestscore{90.0} & 19.9$^{*}$ \\
    \rowcolor{tablebest} \textbf{\method (ours)} & \bestscore{95.1} & \bestscore{84.1} & \bestscore{92.2} & \secondscore{84.0} & \secondscore{79.4} & \secondscore{89.9} & \bestscore{28.1} \\
    \bottomrule
  \end{tabular}
\end{table}

\noindent\textbf{Learning from agent trajectories.}
Rejection-sampling fine-tuning, introduced for reasoning~\cite{zelikman2022star,yuan2023rft}, is the standard recipe for agents~\cite{chen2023fireact}: the agent is fine-tuned on trajectories whose final outcomes pass the task's checks.
It has scaled to general agent tuning~\cite{zeng2024agenttuning,chen2024agentflan}, tool use~\cite{qin2024toolllm}, GUI and web agents~\cite{xu2025agenttrek,sun2025osgenesis}, and software engineering~\cite{pan2025swegym,yang2025swesmith,jain2025r2egym,tao2026swelego,song2026swemaster}.
RL with trajectory-level rewards complements it~\cite{wei2025swerl,golubev2025multiturnswe}, for device-control, web, and GUI agents~\cite{bai2024digirl,qi2025webrl,wang2025uitars2,yang2026openwebrl}, search agents~\cite{jin2025searchr1}, general multi-turn agents~\cite{wang2025ragen,cao2025skyrlagent}, and coding agents~\cite{luo2025deepswe,yang2025kimidev}.
Technical reports of open models scale these recipes: Kimi K2 fine-tunes on synthesized tool-use trajectories before RL with verifiable and self-critique rewards~\cite{kimiteam2025k2}, DeepSeek-V3.2 synthesizes agentic tasks at scale for RL~\cite{deepseekai2025v32}, and other reports describe the agentic post-training of their models~\cite{glm45team2025glm45,minimax2025m1,qwenteam2025qwen3coder,kimiteam2026k25,glm5team2026glm5,cao2026qwen3codernext,chen2026minimaxm2}.
ETO learns from contrasting failed and successful trajectories~\cite{song2024eto}, Agent-R constructs trajectories that recover from errors~\cite{yuan2025agentr}, EEF trains on the beneficial actions of failed expert trajectories~\cite{lan2025eef}, and hindsight relabeling turns trajectories into demonstrations of the goals they actually reach~\cite{li2026hsl}.
Other work refines which steps are imitated.
IPR and STeCa locate steps that fall short of the expert under Monte Carlo step rewards and correct them~\cite{xiong2024ipr,wang2025steca}; SWE-TRACE distills a corpus biased toward shortest-path trajectories~\cite{han2026swetrace}, TQS ranks successful trajectories by structural signals without an LLM judge~\cite{li2026tqs}, and LIMI finds that a small set of curated long-horizon trajectories outperforms large corpora~\cite{xiao2025limi}; WebSTAR keeps only the steps that a judge rates correct~\cite{he2025webstar}, SRFT keeps steps that a critic LLM rates wrong in the context but masks them from the loss~\cite{slinko2026srft}, and SWE-Prime selects trajectory segments by their contribution to the solution and masks the rest from the loss~\cite{sweprime2026}.
These filters read the logged text or the structure of a trajectory; SWE-Prime and SRFT, for example, judge logged steps with an LLM without executing them.
In visual coding, the website can be rendered and tested once it runs, so the effect of a verification step can be measured on the page, including effects that never reach an error log.
Selecting steps also leaves the strategies behind them unnamed; strategy evolution writes them into the playbook (\cref{sec:evolve}).

\noindent\textbf{Beyond outcome supervision: dense credit.}
Process reward models score the steps of a solution with human labels~\cite{lightman2024verify} or Monte Carlo estimates~\cite{wang2024mathshepherd}; for agents, they are trained from rollouts~\cite{choudhury2025agentprm} or from the promise and progress of each step~\cite{xi2025agentprm}, and PRO-CUA uses one for step-level RL of computer-use agents, compared with filtered behavior cloning and trajectory-level RL~\cite{he2026procua}.
Credit assignment for multi-turn RL uses utterance-level critics~\cite{zhou2024archer} and turn-level rewards~\cite{wei2025turnlevel}, and recent methods refine the advantage of each step: GiGPO groups actions taken from the same anchor state to compute step-level advantages~\cite{feng2025gigpo}, SPA-RL trains a progress estimator that decomposes the final reward into stepwise contributions~\cite{wang2025sparl}, ProAS shapes group advantages with a progress-aware reference graph~\cite{ding2026proas}, TRCA scores each state transition with rubrics~\cite{zhang2026trca}, and HCAPO refines step-level values with the LLM as a hindsight critic~\cite{tan2026hcapo}; SWEET-RL instead trains a turn-level critic that has privileged training-time information~\cite{zhou2025sweetrl}.
Closest to our step-level evaluation, the VTCA method of WorkRover scores each turn by the change in deterministic workspace checks and passes a bounded share of positive progress to prerequisite turns through recorded tool-output and artifact dependencies~\cite{he2026workrover}.
Our verification contributions likewise come from check changes (\cref{eq:verify}); in addition, perception steps are credited by their effect on the final score, measured by masking and re-running (\cref{eq:information}), and coding steps by the estimated change in requirement implementation before the website runs (\cref{eq:coding}).
All of these methods credit what the agent does, so RL reinforces a strategy only as often as the agent tries it; self-improvement through RL depends on whether the initial model already exhibits the relevant behaviors~\cite{gandhi2025cognitive}.
Strategy evolution and strategy internalization therefore make effective strategies explicit and common in the training trajectories before RL (\cref{sec:evolve,sec:inject}).

\noindent\textbf{Beyond outcome supervision: natural-language guidance.}
OPRO lets an LLM propose prompts from previously scored ones~\cite{yang2024opro}, GEPA evolves prompts by reflecting on trajectories and keeps a Pareto front of candidates~\cite{agrawal2026gepa}, ExpeL extracts insights from experience~\cite{zhao2024expel}, ACE grows contexts as playbooks through generation, reflection, and curation~\cite{zhang2025ace}, and Training-Free GRPO learns experiential knowledge as a token prior~\cite{cai2025tfgrpo}.
All of them keep the learned text in the context at inference.
Strategy evolution shares their search without parameter updates but accepts a revision only if it raises the final score minus costs of whole trajectories on validation tasks by a confidence margin (\cref{eq:deficit}).
Context distillation trains a model to act without a prompt as it would with it~\cite{snell2022context}, and behavior-conditioned SFT distills a handbook of reasoning behaviors into a model~\cite{didolkar2025metacognitive}.
Learning using privileged information gives the learner information at training time only~\cite{vapnik2009lupi}: P2T uses the reference patch as privileged information to rebuild trajectories that are effective and short~\cite{privproc2026}, and SWEET-RL gives a reference solution to its critic~\cite{zhou2025sweetrl}.
Like context distillation, \method internalizes natural-language guidance.
Its privileged information is an evolved playbook shared by all tasks rather than the reference solution of each task used by P2T and SWEET-RL, and it conditions the policy directly, whereas SWEET-RL gives privileged information to a critic.
The playbook shapes both the training data and the SFT input, and RL without it teaches the agent to choose the strategies on its own (\cref{sec:inject}).

\noindent\textbf{Verification and visual tool use.}
Agents increasingly verify their own work: SmartSnap asks GUI agents to collect snapshot evidence of completion~\cite{smartsnap2025}, and terminal agents verify almost always but detect only part of their errors~\cite{terminalverify2026}.
In multimodal reasoning, ``thinking with images'' trains models to crop and inspect images~\cite{zheng2026deepeyes,toolvision2026}, yet returned observations often have no causal effect on the answer~\cite{illusion2026}, and tool access yields little consistent improvement~\cite{dotools2026}.
The information value (\cref{eq:information}) measures this causal effect for the perception steps of coding agents, as the mean final score of re-runs minus that of re-runs in which all observations of an asset or design region are masked and any prepared file of it is withheld.
Long reasoning chains can also lose track of visual input~\cite{sun2025tvc}.

\section{Limitations}
\label{sec:supp_limitations}

\noindent\textbf{Single training runs.}
Each training configuration is trained once.
Long-horizon RL with 256K-token contexts and rollouts that often exceed a hundred turns is costly, so we do not repeat runs with different random seeds and report no seed variance.
The gain over outcome-based training (12.4 points) is large relative to the largest score change under the hyperparameter variations of \cref{sec:supp_hparams} (1.1 points).

\noindent\textbf{Granularity of step-level feedback.}
\method estimates each action's contribution from the evidence available at its stage.
Some edits, however, are useful mainly because they enable later changes: a preparatory refactor may simplify a subsequent implementation without immediately raising the estimated requirement progress.
Evaluating short sequences of related edits could better capture these delayed contributions.\looseness=-1

\noindent\textbf{Scope of strategy discovery.}
The strategies that \method discovers depend on the tasks and behaviors represented during playbook evolution.
Validation can establish whether a proposed revision improves performance on those tasks, but it does not reveal which useful strategies were never proposed.
A promising direction is to select tasks that expose the agent to situations not yet covered by the playbook.